%% file: report.tex
\RequirePackage{silence}
\documentclass{tudelft-report}
\usepackage[numbers]{natbib}
\usepackage{changes}
\usepackage{amsmath}
\usepackage{amssymb}
\usepackage{relsize}
\usepackage[ruled,vlined]{algorithm2e}
\usepackage{todonotes}
\usepackage{multirow}
\usepackage{adjustbox}
\usepackage{tabularx}
\usepackage{setspace} 
\usepackage{longtable}
\usepackage{amsmath}
\usepackage{subcaption}
\usepackage{caption}
\usepackage{enumitem}
\usepackage{forest}

\usepackage{mdframed}   
\mdfdefinestyle{exampledefault}{%
rightline=true,innerleftmargin=10,innerrightmargin=10,
frametitlerule=true,frametitlerulecolor=green,
frametitlebackgroundcolor=yellow,
frametitlerulewidth=2pt}

\usepackage{adjustbox}
\usepackage{booktabs}

\usepackage{afterpage}

\newcommand{\qedsymbol}{$\blacksquare$}

\newif\ifshowcomments
\showcommentstrue
\ifshowcomments
\newcommand{\mynote}[2]{\fbox{\bfseries\sffamily\scriptsize{#1}}
{\small$\blacktriangleright$\textsf{\emph{#2}}$\blacktriangleleft$}}
\else
\newcommand{\mynote}[2]{}
\fi

\newcommand\blankpage{%
    \null
    \thispagestyle{empty}%
    \addtocounter{page}{-1}%
    \newpage}
\begin{document}

\frontmatter

\input{Content/Front-matter/Cover.tex}

\input{Content/Front-matter/Title.tex}

\input{Content/Front-matter/Abstract.tex}
\input{Content/Front-matter/Preface.tex}

\tableofcontents
\afterpage{\blankpage}

\mainmatter
\input{Content/Chapters/1_Introduction}

\afterpage{\blankpage}
\input{Content/Chapters/2_Preliminary}

\input{Content/Chapters/3_Exploratory_Study}

\input{Content/Chapters/4_CTABGAN}
\input{Content/Chapters/5_DP_CTABGAN}

\afterpage{\blankpage}
\input{Content/Chapters/6_Conclusion}

\appendix
\input{Content/appendix}

\bibliography{report}
\afterpage{\blankpage}

\end{document}

%% file: Content/Front-matter/Cover.tex
\title[tudelft-white]{\Huge Effective and Privacy Preserving \\
Tabular Data Synthesizing}
\author[tudelft-white]{Aditya Kunar}
\affiliation{Technische Universiteit Delft}
\coverimage{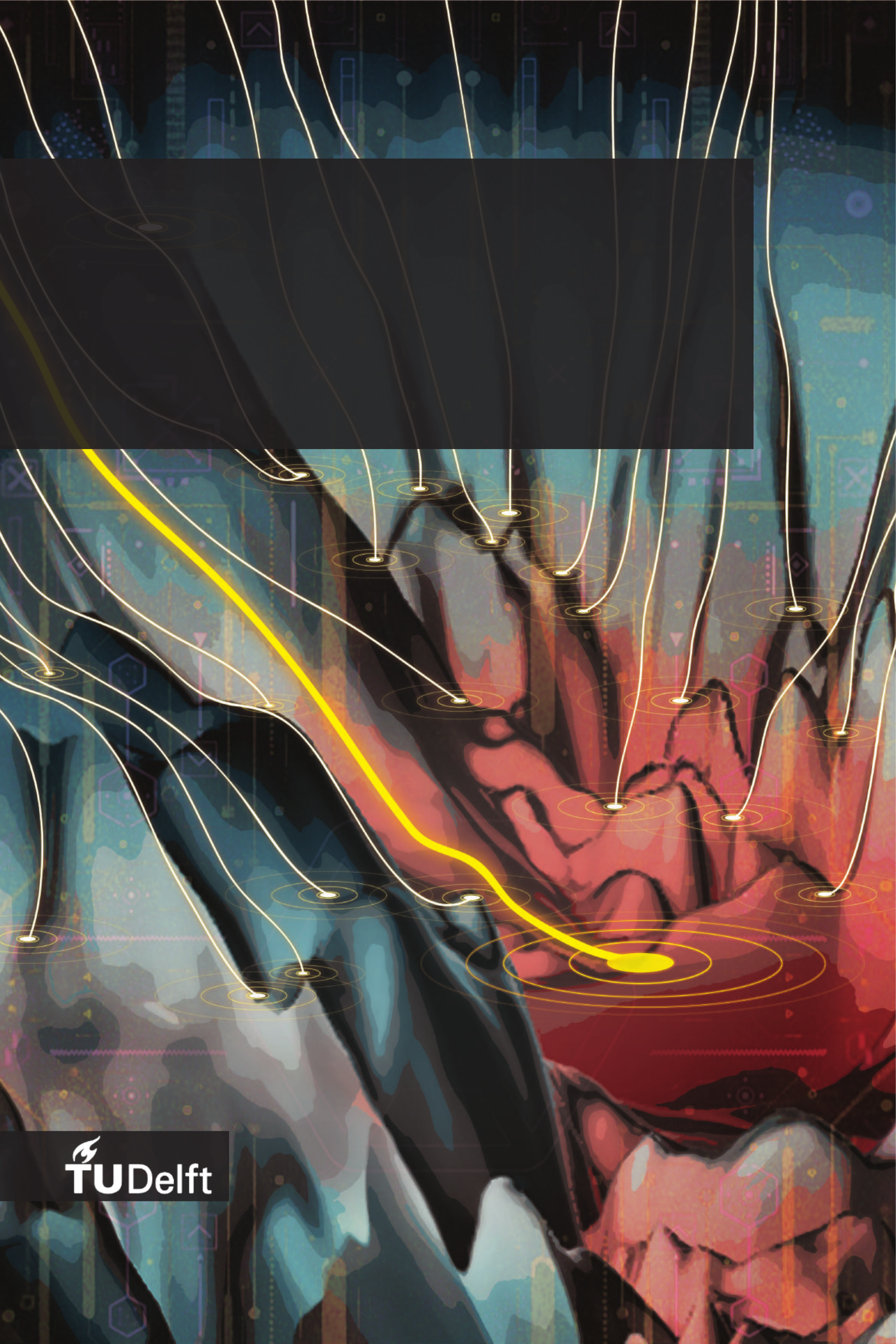}
\titleoffsetx{1.6cm} 
\titleoffsety{10cm}
\afiloffsetx{1cm}
\afiloffsety{18cm}

\makecover

%% file: Content/Front-matter/Title.tex
\begin{titlepage}

\begin{center}


{\makeatletter
\largetitlestyle\fontsize{64}{94}\selectfont\@title
\makeatother}

{\makeatletter
\ifx\@subtitle\undefined\else
    \bigskip
   {\tudsffamily\fontsize{22}{32}\selectfont\@subtitle}    
\fi
\makeatother}

\bigskip
\bigskip

by

\bigskip
\bigskip

{\makeatletter
\largetitlestyle\fontsize{26}{26}\selectfont\@author
\makeatother}

\bigskip
\bigskip

to obtain the degree of Master of Science

at the Delft University of Technology,

to be defended publicly on Friday July 9, 2021 at 10:00 AM.

\vfill

\begin{tabular}{lll}
    Student number: & 5074274 \\
    Project duration: & \multicolumn{2}{l}{November 1, 2020 -- July 09, 2021} \\
    Thesis committee: & Dr.\ Lydia\ Chen, & TU Delft, supervisor \\
        & Dr.\ Jan\ van Gemert, & TU Delft \\
        & Mr.\ Hiek\ Scheer, & Aegon \\
        & Mr.\ James\ Gnanasekaran, & Aegon \\
        
\end{tabular}

\bigskip
\bigskip
\emph{This thesis is confidential and cannot be made public until August 31, 2021.}

\bigskip
\bigskip
An electronic version of this thesis is available at \url{http://repository.tudelft.nl/}.


\end{center}

\begin{tikzpicture}[remember picture, overlay]
    \node at (current page.south)[anchor=south,inner sep=0pt]{
        \includegraphics{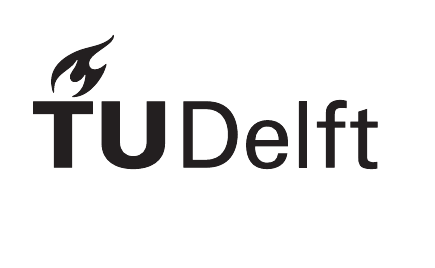}
    };
\end{tikzpicture}

\end{titlepage}

%% file: Content/Front-matter/Abstract.tex
\chapter*{Abstract}
\addcontentsline{toc}{chapter}{Abstract}
\setheader{Abstract}
\vspace*{1.5cm}
\begin{quote}
    While data sharing is crucial for knowledge development, privacy concerns and strict regulation (e.g., European General Data Protection Regulation (GDPR)) unfortunately limit its full effectiveness. Synthetic tabular data emerges as an alternative to enable data sharing while fulfilling regulatory and privacy constraints. The state-of-the-art tabular data synthesizers draw methodologies from Generative Adversarial Networks (GAN). In this thesis, we develop CTAB-GAN, a novel conditional table GAN architecture that can effectively model diverse data types with complex distributions. CTAB-GAN is extensively evaluated with the state of the art GANs that generate synthetic tables, in terms of data similarity and analysis utility. The results on five datasets show that the synthetic data of CTAB-GAN remarkably resembles the real data for all three types of variables and results into higher accuracy for five machine learning algorithms, by up to 17\%. 

    Additionally, to ensure greater security for training tabular GANs against malicious privacy attacks, differential privacy (DP) is studied and used to train CTAB-GAN with strict privacy guarantees. DP-CTAB-GAN is rigorously evaluated using state-of-the-art DP-tabular GANs in terms of data utility and privacy robustness against membership and attribute inference attacks. Our results on three datasets indicate that strict theoretical differential privacy guarantees come only after severely affecting data utility. However, it is shown empirically that these guarantees help provide a stronger defence against privacy attacks. Overall, it is found that DP-CTABGAN is capable of being robust to privacy attacks while maintaining the highest data utility as compared to prior work, by up to 18\% in terms of the average precision score.   
    \begin{flushright}
    {\makeatletter\itshape
        \@author \\
        Delft, August 2021
    \makeatother}
    \end{flushright}
\end{quote}




%% file: Content/Front-matter/Preface.tex
\chapter*{Preface}
\addcontentsline{toc}{chapter}{Preface}

\setheader{Preface}
\vspace*{1.5cm}
\begin{quote}

My thesis builds upon a background in adversarial machine learning with a focus on tabular data generation and it's use as both an imminent and eminent privacy preserving technology. The research begins by attempting to improve state-of-art GAN based tabular data synthesizers by understanding their fundamental strengths and weaknesses in~\autoref{ch3}. Based on the conclusions drawn from this study existing weaknesses are addressed and the strengths of various methodologies are combined to enhance performance, giving rise to a novel tabular data generator, CTAB-GAN, in~\autoref{ch4}. Additionally, privacy concerns for synthetic tabular data generation are addressed by studying the use of differential privacy. And, an empirical investigation of privacy exposure is carried out using membership and attribute inference based attacks in~\autoref{ch5}. \\

This research is a product of the wonderful guidance of my supervisor Dr. Lydia Chen and my daily supervisor Dr. Zilong Zhao. I owe a great sense of gratitude to both of them for their extensive knowledge which they willfully provided throughout the period of my master’s thesis. Finally, I offer my thanks to my family and friends who have been a pillar of support.\\

I would also like to express my gratitude towards Mr. Hiek Scheer and Mr. James Gnanasekaran, who graciously accepted to be a part of the defence committee.\\

\begin{flushright}
{\makeatletter\itshape
    \@author \\
    Delft, August 2020
\makeatother}
\end{flushright}

\end{quote}

%% file: Content/Chapters/1_Introduction.tex
\chapter{Introduction}\label{ch1}

“Data is the new oil” is a quote that goes back to 2006, which is credited to mathematician Clive Humby. It has recently picked up more steam after The Economist published a 2017 report~\cite{theeconomist} titled “The world’s most valuable resource is no longer oil, but data”. This thesis focuses on tabular data, the most popular type of data to be used for analysis in the industry~\cite{sun2019supertml}.

Unfortunately, extracting insights from tabular data risks losing personal privacy and results in an unjustified analysis~\cite{narayanan2008}. Thus, strict privacy regulations enforced via the European General Data Protection Regulation (GDPR) prevent the misuse of personal data. This calls for innovative technologies that can enable data-usage without breaching privacy. Hence, privacy preserving data solutions have become increasingly important and have the potential to push the contribution of the data economy to the EU GDP by up to 4\%~\cite{data-sharing}.

One such emerging solution is to leverage Generative Adversarial Networks (GAN)~\cite{gan}. GANs are first trained on a real dataset and are then subsequently used to generate synthetic data resembling the original data distribution. Beyond successfully generating images, GANs have recently been applied to generate high quality tabular datasets~\cite{ctgan, tablegan}. And since generative model can synthesize fake data as many as we want, it is good when available data is limited. For example, the case for online learning \cite{rad}.

Currently, the state-of-the-art tabular generators~\cite{ctgan} use the conditional GAN architecture and deal with only on two types of variables, namely continuous and categorical. However, an important class of "mixed" data-types is overlooked. In addition, existing solutions cannot effectively handle highly skewed continuous variables. And finally, the empirical robustness of existing methods to withstand malicious privacy attacks remains unexplored.

In this thesis, we design a tabular data synthesizer that addresses the limitations of the prior work by: (i) efficiently encoding "mixed" data-types consisting of both continuous and categorical variables, (ii) efficiently modeling skewed continuous variables and (iii) enhancing robustness against privacy attacks. Therefore, we propose a novel conditional tabular generative adversarial network, CTAB-GAN, that is further extended to be trained with strict privacy guarantees.

Thus, this chapter begins with Sec.~\ref{Ch1:Syn} elaborating on the necessity of synthetic data with robust privacy guarantees along with beneficial use-cases in the industry. Next, Sec.~\ref{Ch1:motivation} specifies the key scientific motivations for this research. This is followed by the main research questions posed in Sec.~\ref{Ch1:research_question} along with the main results, collaborations and contributions of this thesis in Sec.~\ref{Ch1:Res} \& Sec.~\ref{Ch1:contri}, respectively. Finally, Sec.~\ref{Ch1:org} ends with an outline of how this research is organised.

\section{Privacy Preserving Synthetic Tabular Data}
\label{Ch1:Syn}

Tabular data plays a key role in a wide-range of industries for gaining valuable insights and making data-driven decisions. For e.g., consider the recommendation systems employed on our favourite websites such as Netflix or Bol.com. Or, the corona patient risk models developed by our health-care providers. These all intimately rely on tabular data. 

But unfortunately, using the real tabular data may be perilous because: (i) the \textit{privacy} of real data may be comprised (ii) the \textit{standard} of real data may be poor due to rows with incomplete information and (iii) the \textit{amount} of real data representing anomalous events (e.g., data-rows representing "fraud") may be heavily imbalanced as compared to normal events (e.g., data-rows representing "no fraud"). 

These factors necessitate the use of synthetic tabular data to ensure that the data doesn't contain any real user-sensitive information compromising \textit{privacy}, missing values degrading \textit{quality} and contains a balanced \textit{quantity} of class labels (e.g., equal number of data-rows representing "fraud" vs "no fraud", respectively).

Additionally, due to the recent rise in machine learning solutions that rely on real user-data, there has been an equally important demand for ensuring greater data security against malicious privacy attacks targeted towards machine learning algorithms. 

In light of this, privacy-preserving techniques such as differential privacy\cite{dwork2008differential} serve as an effective framework to limit the influence of individual data points and to provide strict privacy guarantees preventing the loss of personal information. In recent times, tech giants such as Apple\cite{tang2017privacy} have successfully used this technique to effectively deal with privacy leaks.  \\
\\
Thus, synthetic tabular data generated using strict differential privacy guarantees serves data-driven industries with the following gains:
\begin{itemize}
    \item \textbf{Collaboration across stakeholders}- Synthetic data with reliable privacy guarantees serves to enable efficient and safe data-disclosure. This boosts collaboration among different parties and fosters innovation. For e.g., for building a stronger fraudulent insurance claim detector, a multi-national insurance company can benefit from information stored between divisions located across the world. However, privacy restrictions do not allow the real data to be shared. Thus, synthetic data can be used instead, to capture the shared characteristics of fraudulent insurance claims across the world.  
    
    \item \textbf{Data Optimization}- Synthetic data generators can effectively learn the distribution of the real data thereby enabling end-users to encapsulate the real information in a more compressed form. This enables easily storing and generating large amounts of data more efficiently. Moreover, synthetic data generators can generate datasets based on user-specified constraints and do not contain missing values by design~\cite{ctgan}.
    
    \item \textbf{Model Optimization}- To improve the performance of machine learning algorithms, synthetic data can be used for performing data-augmentation to effectively re-balance datasets with imbalanced class labels~\cite{engelmann2020conditional}. Moreover, the synthetic data can be used as a proxy validation-dataset to tweak and validate the most optimal hyper-parameters thereby allowing a more efficient usage of the real data for training machine learning models~\cite{fintz2021synthetic}. 
\end{itemize}

\section{Motivation}\label{Ch1:motivation}

The industrial datasets (at stakeholders like banks, insurance companies, and health care) present multi-fold challenges. First of all, such datasets are organized in tables and populated with both continuous and categorical variables, or a mix of the two, e.g., missing values can be considered to be categorical elements embedded in continuous variables as they are clearly separate from the continuous variable's original distribution. Here, such type of variables are termed as "mixed" variables. Secondly numeric data variables often have a wide range of values as well as a skewed distributions, e.g., the statistic of the transaction amount for a credit card. Most transactions should be within 0 and 500 bucks (i.e. daily shopping for food and clothes), but exceptions of a high transaction amount surely exist. And last but not least, training tabular GANs with sensitive datasets risks leaking privacy through malicious privacy attacks.
\\
\\
In summary, dealing with the following challenges formed the main motivations of research:

\begin{itemize}
    
    \item Tabular data comprises of "mixed" variables that consist of both a continuous and a categorical component. 
    
    \item Continuous variables exhibit heavily skewed distributions which are difficult to model and reproduce authentically.
    
    \item Tabular GANs comprise the privacy of the original dataset used for training.
\end{itemize}

\section{Research Questions}\label{Ch1:research_question}

Building on the motivations established for the thesis in ~Sec.~\ref{Ch1:motivation}, the thesis revolves around three main research question as follows: 

\begin{itemize}
    \item What are the performance capabilities of existing tabular GANs?
    \item How to improve upon the tabular generation quality of state-of-the-art tabular GANs?
    \item How to train tabular GANs in a privacy-preserving manner?
\end{itemize}

The main research questions are then further divided into the following sub-questions: 
\begin{enumerate}
    \item What are the performance capabilities of existing tabular GANs?
    \begin{enumerate}
        \item What is the statistical similarity, ML utility and privacy risk concerning synthetically produced datasets in terms of their corresponding original datasets?
        \item What are the challenges faced by existing tabular GANs?
    \end{enumerate}
    \item How to improve upon the tabular generation quality of state-of-the-art tabular GANs?
    \begin{enumerate}
        \item How to handle "mixed" variables in tabular data?
        \item How to deal with skewed continuous variables?
    \end{enumerate}
    \item How to prevent privacy leakage for tabular GANs?
    \begin{enumerate}
        \item How can differential privacy guarantees be instilled for synthetic tabular data generation?
        \item Do the theoretical privacy guarantees successfully prevent privacy leakage?
    \end{enumerate}
\end{enumerate}

\section{Main Results and Collaborations}
\label{Ch1:Res}		
\subsection{Publications}
 The work developed in this thesis has lead to several contributions which have been submitted in various venues:

 \begin{itemize}
 
  	\item Aditya Kunar, Robert Birke, Zilong Zhao, Lydia Y. Chen. \textbf{DTGAN: Differential Private Training for Tabular GANs.}, under review \cite{dtgan}.
 
  	\item Zilong Zhao, Aditya Kunar, Robert Birke, Lydia Chen. \textbf{CTAB-GAN: Effective Table Data Synthesizing}, under review \cite{ctabgan}. \vspace{-0.5em}
  	
 	\item Zilong Zhao, Aditya Kunar, Robert Birke, Lydia Chen. \textbf{FedTGAN: Federated Learning Framework for Synthesizing Tabular Data}, under review. \vspace{-0.5em}

 \end{itemize}

 		\subsection{Collaborations}
 Those works has been conducted thanks to fruitful collaborations:
 \begin{itemize}    
   \item Dr. Lydia Y. Chen (TU Delft) on tabular GAN algorithm, differential privacy and distributed GAN algorithm,\vspace{-0.5em} 
   \item Dr. Robert Birke (ABB Research) on tabular GAN algorithm, differential privacy and distributed GAN algorithm, \vspace{-0.5em}
   \item Dr. Zilong Zhao (TU Delft) on tabular GAN algorithm, differential privacy and distributed GAN algorithm,\vspace{-0.5em} 
 \end{itemize}

\section{Contribution of thesis}\label{Ch1:contri}

Our primary contributions of this thesis have been the following:

\begin{itemize}
 
\item An extensive bench-marking of 4 state-of-the-art tabular GANs in terms of statistical similarity, ML utility and privacy. And, emphasizing important issues faced by existing methods. 

\item A novel conditional generative adversarial network, CTAB-GAN, that can effectively handle "mixed" data-types and skewed continuous variables. 
\item Differential private training of CTAB-GAN and rigorous privacy risk evaluation against membership and attribute inference attacks.   

\end{itemize}

\section{Report Outline}\label{Ch1:org}
The thesis has the following outline, in~\autoref{ch2}, the relevant related work and core concepts pertaining to generating privacy-preserving synthetic data is highlighted. In~\autoref{ch3}, an exploratory study quantitatively evaluating 4 state-of-the-art tabular GAN approaches is elucidated. Moreover, the chapter highlights challenges faced by existing methods. In~\autoref{ch4}, a novel conditional table generative adversarial network, CTAB-GAN, is proposed to improve on challenges faced by the state-of-the-art. In~\autoref{ch5}, the application of differential privacy in the context of tabular GANs is examined and the empirical robustness against privacy attacks is studied. Lastly, in~\autoref{ch6}, we finally summarise this thesis by reviewing the research questions established in this chapter and by identifying limitations of CTAB-GAN and defining avenues of research for future work.

%% file: Content/Chapters/2_Preliminary.tex
 \chapter{Related Work $\&$ Preliminaries}\label{ch2} 

This chapter begins with Sec.~\ref{Ch2:Related_work} discussing the relevant literature for tabular GANs and their differential private variants. And, Sec.~\ref{Ch2:TC} provides a brief primer on generative adversarial networks (GANs) and differential privacy (DP) in the context of tabular data.

\section{Related Work}
\label{Ch2:Related_work}

\resizebox{\columnwidth}{!}{
\begin{forest}
     for tree={
      align=center,
      edge+={thick},
      draw,
      fill=pink!60,
      rounded corners=2pt,
      drop shadow,
    },
    if level=0{
      tikz={\draw [thick] (.children first) (.children last);}
    }{},
    [\textbf{GAN}
    [
    Tabular GAN\\~[Non-conditional GAN\\~[MedGAN\cite{choi2017generating}][TableGAN\cite{tablegan}]]
    [Conditional GAN\\~[CTGAN\cite{ctgan}]
    [CW-GAN\cite{engelmann2020conditional}]]
    ]
    [DP-GAN\\~[PATE\cite{papernot2016semi}\\~[PATE-GAN\cite{pategan}]]
    [DP-SGD\cite{abadi2016deep}\\~[Generator\\~[GS-WGAN\cite{chen2020gs}]]
    [Discriminator\\~[DP-WGAN\cite{xie2018differentially}]]
    ]
    ]
        ]
      ]
    ]
  \end{forest}
}

\subsection{Tabular GANs}
\label{Ch3:SOTA}

In this section, the focus is on GAN-based methods that deal with tabular data generation. These methods are featured extensively in the work done in \autoref{ch3} and \autoref{ch4}. Tab.~\ref{table:sota} details key features for each method.\\
\\
\textbf{MedGAN}- In the work done by \cite{choi2017generating} (2017), the authors propose a novel mechanism to synthetically generate Electronic Health Records (EHR) consisting of high dimensional categorical variables. Their model consists of using a combination of an auto-encoder and a generative adversarial network. They show that their model is capable of producing realistic synthetic patient records as evaluated via a qualitative medical expert review. Additionally, they empirically analyze the risk of violating privacy via identity and attribute disclosure attacks and conclude that the risk is manageable. 

However, the MedGAN model cannot generate synthetic datasets outside the medical domain of Electronic Health Records which contain only categorical variables.\\
\\
\textbf{TableGAN}- In the work done by \cite{tablegan} (2018), the authors develop a tabular data synthesizer that is based on the \textit{DCGAN architecture} (refer to Sec.~\ref{Ch4:dcgan}). Their approach utilizes a separate classifier module in addition to the discriminator and generator modules commonly used in GAN-based frameworks. Moreover, their method relies on additional loss objective for the generator known as the \textit{classification} $\&$ \textit{information losses}, respectively (refer to Sec.~\ref{Ch4:tabloss}). 

However, TableGAN doesn't deal with generating categorical variables in a principled manner. This is because their approach involves mapping categorical variables to integers and treating them purely numerically.\\
\\
\textbf{CTGAN}- In the work done by \cite{ctgan} (2019), they introduce a novel conditional tabular GAN architecture as well as the training-by-sampling method (refer to Sec.~\ref{Ch4:cgan}). These improvements allow the generator to more efficiently produce realistic samples for the minority categories found in discrete columns thereby producing synthetic records which match the real data distribution more closely. In addition, they introduce the mode-specific-normalization technique (refer to Sec.~\ref{Ch4:msn}) for learning complex numerical distributions. Lastly, their discriminator network is trained with WGAN loss with gradient penalty~\cite{gulrajani2017improved} (refer to Sec.~\ref{Ch5:WGAN}) for improved training of GANs. 

However, the CTGAN model is incapable of dealing with missing values. This limits it's applicability in real-world scenarios where the data is often impure and contains a large number of missing values. Moreover, the authors also convey that learning from small training sets severely affects performance as well. \\ 
\\    
\textbf{Conditional Wasserstein GAN}- In the work done by \cite{engelmann2020conditional} (2020), they propose the conditional Wasserstein GAN primarily for the purposes of data augmentation. Similar to TableGAN, they exploit the classification loss and augment it to the generator's loss objective. Moreover, their model also integrates cross-layers~\cite{wang2017deep} in both the discriminator and generator networks. 

However, in their work, they do not make use of any activation functions for generating samples for numerical attributes. This makes it inherently difficult to constrain the generator's output to a meaningful range of values. Moreover, it can also lead to severely destabilizing the training process.

\begin{table}[htb]
\centering
\caption[BB]{\centering State-of-the-Art Blueprint\footnotemark.}
\resizebox{0.8\columnwidth}{!}{
\begin{tabular}{|c|c|c|c|c|}
\hline
\textbf{Method} & \textbf{Data format} & \textbf{Training method} & \textbf{Privacy analysis} & \textbf{Designated output} \\
\hline
MedGAN    &  Categorical Only  &  Auto-encoder + GAN            & Yes & No   \\
TableGAN  &  Categorical $\&$ Continuous     &  GAN + Classifier  & Yes & No   \\
CTGAN     &  Categorical $\&$ Continuous     &  Conditional-WGAN  & No  & Yes  \\
CW-GAN    &  Categorical $\&$ Continuous     &  Conditional-WGAN + Classifier & No  & Yes  \\
\hline
\end{tabular}
}
\label{table:sota}
\end{table}
\footnotetext{ Note that the designated output column is used to identify models which have the capacity to sample data-instances with user-defined categorical attributes.
}

\subsection{Differential Private GANs}
\label{Ch5:related_work}

In this section, relevant differential private GAN models are reviewed in relation to the work done in \autoref{ch5}. Tab.~\ref{Ch5:tab1} highlights key details of each model. \\
\\
\textbf{PATE-GAN}- In the work done by \cite{pategan} (2019), the authors devise a technique for integrating DP guarantees in tabular GANs via the Private Aggregation of Teacher Ensembles (PATE) framework~\cite{papernot2016semi}. In their approach, multiple teacher discriminators are trained using disjoint subsets of the training data along with a student discriminator where the aggregation of the teacher ensemble is done after perturbing the predictions of teacher discriminators using Laplacian noise. 

However, PATE-GAN suffers from the following limitations: (i) the student discriminator is trained solely using generated samples and does not see any real samples. This is problematic because if the student discriminator only has access to the unrealistic samples generated by the generator, it won't provide reliable feedback to the generator so that it can improve it's sample quality (refer to Sec.~\ref{Ch5:Res} for experimental evidence) and (ii) The PATE-GAN framework requires careful hyper-parameter tuning to select the number of teacher discriminators. \\
\\    
\textbf{DP-WGAN}- In the work done by \cite{xie2018differentially} (2018), the authors incorporate differential privacy guarantees for wasserstein GANs wherein DP-SGD\cite{abadi2016deep} is applied for training the discriminator. Moreover, they make use of weight clipping to enforce the Lipschitz constraint on the discriminator so as to be compatible with the wasserstein loss. 

However, the challenges with this approach lies in the fact that calibrating the DP specific hyper-parameters (i.e gradient norm clipping value) varies drastically based on differences in network architectures and training procedures. Additionally, clipping the weights of the discriminator has been found to cause convergence issues\cite{gulrajani2017improved}. \\
\\
\textbf{GS-WGAN}- In the work done by \cite{chen2020gs} (2020), the authors work towards training image GANs in a privacy preserving manner. Their novel contributions include utilising the wasserstein loss with gradient penalty\cite{gulrajani2017improved} to avoid hyper-parameter tuning of the clipping parameter and employing DP guarantees using DP-SGD\cite{abadi2016deep} for the generator network rather than the discriminator network. This is motivated by the fact that only the generator network is made publicly available after training the GAN model. Moreover, they propose a more precise approach for perturbing gradients of the generator by only manipulating those that are computed with respect to the training data so as to minimize the loss of gradient information. And lastly, they make use of the subsampled R\'enyi Differential Privacy (RDP) Accountant\cite{wang2019subsampled} to compute the privacy loss during training. 

However, their work focuses only on DP training of the generator and offers no experimental analysis against privacy attacks.

\begin{table}[htb]
\centering
\caption{\centering Outline of all methodologies in this work.}
\resizebox{0.8\columnwidth}{!}{
\begin{tabular}{ |c|c|c|c|c|c|c|c| }
\hline
\textbf{Model} & \textbf{Loss} &\textbf{DP Site} & \textbf{Noise}& \textbf{Accountant} & \textbf{Data Format}\\ 
\hline
PATE-GAN & KL Divergence Loss  &  Discriminator   & Laplacian  & PATE Accountant & Table\\
DP-WGAN & Wasserstein Loss + Weight Clipping    &  Discriminator   & Gaussian  & RDP-Accountant & Image $\&$ Table \\ 
D-DP-CTABGAN & Wasserstein Loss + Gradient Penalty & Discriminator  & Gaussian & RDP-Accountant & Table\\ 
G-DP-CTABGAN & Wasserstein Loss + Gradient Penalty & Generator  & Gaussian & RDP-Accountant & Table\\ 
\hline
\end{tabular}
}
\label{Ch5:tab1}
\end{table}

\section{Background}
\label{Ch2:TC}
\subsection{GAN Designs}
\label{Ch4:TB}
\textbf{GAN}- \textbf{GANs} are a relatively recent breakthrough in machine learning and generative modelling. Unlike conventional machine learning algorithms that learn a conditional distribution of a class variable given the predictor variables (e.g., to solve a binary classification problem), the main purpose of GANs is to learn the joint distribution of the entire input data. In this way, using the learned joint distribution, the generated samples can be drawn to resemble the original input data. 

GANs make use of two neural networks: the generator and the discriminator networks. The generator takes as input a random noise vector to synthesize data that closely resembles the real data. Whereas, the discriminator takes as input real/generated samples and acts a teacher assessing the output of the generator judging whether the generated samples are real or fake. Much like a supervisor providing feedback to a student about his/her work. The two models are trained together via an adversarial min-max game minimizing the loss of the generator while maximizing the loss of the discriminator expressed below~\cite{gan}: 
\begin{equation}
\label{eq:gan}
\min_{\mathcal{G}}\max_{\mathcal{D}}V(\mathcal{G},\mathcal{D}) = \mathbb{E}[log\mathcal{D}(x)]_{x\sim p_{data}(x)} + \mathbb{E}[log(1-\mathcal{D}(\mathcal{G}(z)))]_{z \sim p(z)}
\end{equation}
where $\mathcal{G}$ $\&$ $\mathcal{D}$ represent the generator and discriminator networks, respectively. Furthermore, $p_{data}$ denotes the real data distribution and $p(z)$ denotes a prior distribution (i.e $\mathcal{N}(0,I)$) with latent vector $z$. And, $\mathcal{D}$ outputs a scalar in the range [0,1].

\begin{figure}[htb]
    \centering
    \includegraphics[scale=.20]{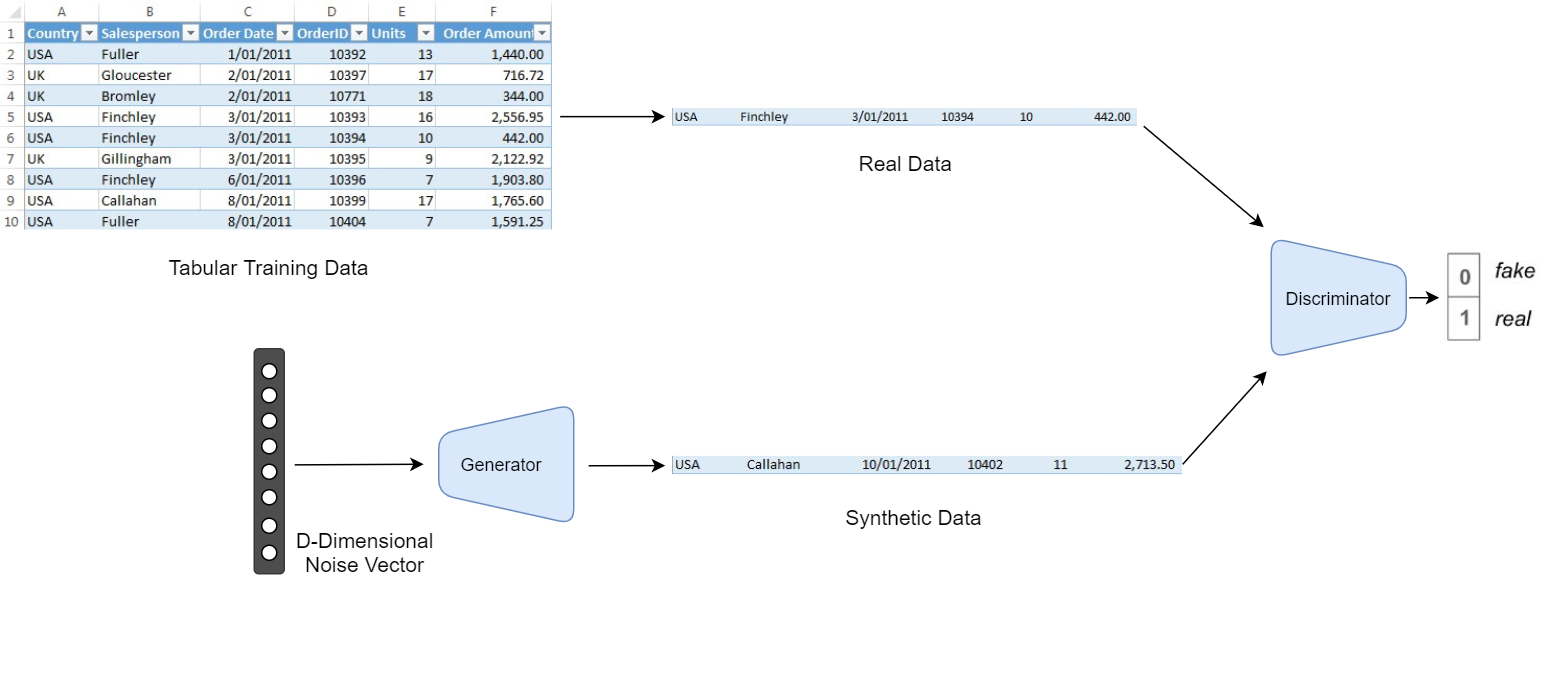}
    \caption{\centering Synthetic Tabular Data Generation via GANs}
    \label{fig:STD}
\end{figure}

\textbf{Tabular GANs} are simply GANs that are used to generate tabular formatted datasets. As an example, consider an SQL table used to store employee information. In this setting, each entry in the table is an independent sample obtained from the joint distribution of all the employees. The goal of tabular GANs is to learn such a joint distribution to subsequently synthesize data that matches the original. Fig.~\ref{fig:STD} illustrates this process. 
\\
\textbf{DCGAN\cite{radford2015unsupervised}}-\label{Ch4:dcgan} The \textbf{DCGAN architecture} is an extension of the standard GAN architecture which makes use of \textit{convolutional} and \textit{convolutional-transpose layers} in the discriminator and generator, respectively. It is a widely used stable GAN architecture that has proven to be useful for generating images as well as tabular data~\cite{tablegan}. 

The generator network of \textit{DCGAN} consists of stacks of \textit{strided 2D convolutional transpose layers} followed by a \textit{2d batch norm layer} and a \textit{ReLU activation function}. The final output of the generator is passed through a \textit{Tanh activation function} to bring the values in the original range of $[-1,1]$ (for representing images). The generator takes as input a random noise vector of arbitrary length and returns an image with the same spatial dimensions as the original dataset. 

Whereas, the discriminator network is composed of stacks of \textit{2d convolutional layers}, \textit{2d batch norm} and \textit{LeakyReLU layers} with a leaky ratio of $0.20$ with the final output being passed through a \textit{sigmoid activation function}. The discriminator takes as input real/fake images and outputs the probability of any particular sample being real or synthetic. 

Lastly, it is worth noting that the presence of \textit{batch normalisation} in both the generator and discriminator networks is a key contribution of the authors that leads to a stable flow of gradients for training \textit{DCGAN} reliably. \\
\\
\textbf{Conditional GAN $\&$ Training-by-Sampling\cite{ctgan}}-\label{Ch4:cgan} To address the problem of imbalanced categorical variables in real-world datasets, the \textit{conditional generator}, \textit{generator loss} and \textit{training-by-sampling} are introduced in the work of \cite{ctgan}. The main idea behind these techniques stems from the use an additional vector, termed as the \textit{conditional vector}, to represent the classes of categorical variables. This vector is both fed to the generator and used to bound the sampling of the real training samples to subsets satisfying the condition. Moreover, the conditions are sampled in such a way so as to give higher chances to minority classes while training the model. These concepts are explained in greater detail below. \\
\\
The \textbf{Conditional GAN} features a \textbf{conditional generator} whose generated samples come from a \textit{conditional probability distribution} $\hat{r} \sim \mathbb{P}_{\mathcal{G}}(row|C_{i^{*}}=c^{*})$ where, $c^{*}$ is a particular class within the $i^{th}$ categorical variable
$C_{i^{*}}$. Intuitively, this corresponds to generating a row given a chosen class for a selected categorical variable. To represent this condition (i.e., $C_{i^{*}}=c^{*}$), the \textit{conditional vector} is used. 

To construct the \textit{conditional vector}, \cite{ctgan} treats all categorical variables $C_{1},...,C_{N_c}$ as one-hot vectors $c_1,...,c_{N_c}$ where $N_c$ represents the total number of categorical variables. Let the $i^{th}$ one-hot vector and it's corresponding $mask$ vector be denoted as $c_i=[c_i^{(k)}]$, for $k = 1,...,|C_i|$ $\&$ $m_{i}=[m_i^{(k)}]$, for $k = 1,...,|C_i|$, respectively. The condition is then expressed using the $masks$ for each one-hot vector as: $m_i^{(k)} = \{1, $ if $ i=i^{*}$ $\&$ $k=k^{*}$, else $0\}$ where $i^{*}$ is a chosen categorical variable and $k^{*}$ is the selected class within variable $i^{*}$. Thus, the \textit{conditional vector} is represented as: $cond = m_{1}\oplus...\oplus m_{N_{c}}$ where $\oplus$ is the concatenation operator. As an example, consider two categorical variables $C_1= [0,1]$ and $C_2=[0,1]$, if the condition is $C_1=0$, the corresponding $masks$ will be $m_1 = [1,0]$ $\&$ $m_2=[0,0]$ to result in a \textit{conditional vector} i.e., $cond=[1,0,0,0]$. \\
\\
Next, \cite{ctgan} uses the \textbf{generator loss} to ensure that the \textit{conditional generator} generates samples that match the constraint provided by the \textit{conditional vector}. As an example, consider the $mask$ $m_i = [1,0]$ for a particular categorical variable $i$. Given this condition, the \textit{conditional generator} should ideally similarly output a data row where the $0^{th}$ class for the $i^{th}$ categorical variable is produced leading to a matching generated $mask$ $\hat{m}_{i} = [1,0]$. Thus, if $m_{i}$ represents the conditional $mask$ for the selected one-hot-encoded variable $i$ associated for a given data row and $\hat{m}_{i}$ denotes the corresponding generated $mask$, the \textit{generator loss} denoted as $\mathcal{L}_{generator}^{G}$ is formally represented as: $H(m_{i}, \hat{m}_{i})$ where $H(.)$ is the \textit{cross-entropy loss}. Therefore, in this manner, the added loss acts as a soft constraint for enforcing that the generated samples are aligned with their corresponding \textit{conditional vectors.}\\
\\
Finally, the \textbf{training-by-sampling} method is used to sample the \textit{conditional vector} in such a way so that the model can explore all possible classes present in categorical variables evenly during training. Thus, the sampling procedure for generating a condition is as follows:
\begin{enumerate}
    \item Out of $N_c$ categorical variables, a column $C_{i^{*}}$ is uniformly chosen at random with probability $1/N_{c}$.
    \item Based on the chosen column $C_{i^{*}}$, a probability mass function (PMF) is created after applying a \textit{log-transformation} to the frequency of individual classes within column $C_{i^{*}}$ where the log-transform naturally leads to an over-sampling of minority classes.
    \item On the basis of the constructed PMF described above, a class $c^{*}$ is sampled for the selected column $C_{i^{*}}$. Thus the condition $C_{i^{*}}=c^{*}$ and it's corresponding \textit{conditional vector} $cond$ is formed.  
\end{enumerate}

\subsection{GAN Loss Objectives}

\textbf{Wasserstein Loss with Gradient Penalty\cite{gulrajani2017improved}}-\label{Ch5:WGAN} The \textbf{wasserstein loss} first proposed in the work of \cite{pmlr-v70-arjovsky17a} (2017) provides greater stability for training GANs as compared to the classical KL divergence loss (as shown in Eq.~\ref{eq:gan}). In contrast to the KL divergence loss, the proposed loss function remains continuous and differentiable for measuring the similarity of probability distributions with non-overlapping support. Therefore, it is capable of providing more meaningful gradients for training the generator especially for cases where the probability distribution of generated samples is highly dissimilar to the real probability distribution. Formally, the wasserstein loss may be expressed as minimizing the integral probability metrics (IPMs)
$sup_{f\in\mathcal{F}}|\int_{M}fd\mathbb{P}_r - \int_{M}fd\mathbb{P}_g|$ between real($\mathbb{P}_r$) and generated ($\mathbb{P}_g$) data distributions, where $\mathcal{F}=\{f:||f||_{L}\leq 1\}$ enforces the discriminator function $f$ to be 1-Lipschitz continuous. In practice, \cite{pmlr-v70-arjovsky17a} proposed weight clipping to enforce the Lipschitz constraint on the discriminator by clamping the weights of the discriminator to lie within a compact space $[-c,c]$ where $c$ is the clipping threshold. 

However, \cite{gulrajani2017improved} proposed the \textbf{gradient penalty term} as an alternative to weight clipping. As they found that weight clipping may lead to convergence issues by biasing the discriminator towards simpler functions or causing exploding/vanishing gradients. Motivated by their theoretical proof illustrating that an optimal discriminator naturally possesses a gradient norm of 1 almost everywhere under real and generated distributions, $\mathbb{P}_r$ and $\mathbb{P}_g$ respectively, the authors define the discriminator to be 1-Lipschitz continuous if and only if, it has gradients with norm at most 1 everywhere. They then enforce 1-Lipschitz continuity of the discriminator by adding a soft constraint during training to constrain the gradient norm of the discriminator's output with respect to it's input. Originally the authors define the input (i.e random samples $\hat{x}\sim\mathbb{P}_{\hat{x}}$) as sampling along the straight lines between pair of points sampled from the original data distribution $\mathbb{P}_{r}$ and the generator distribution $\mathbb{P}_{g}$. Thus, the training objectives $\mathcal{L}_D$ $\&$ $\mathcal{L}_G$ for the discriminator and generator are expressed as:
\begin{equation}
    \mathcal{L}_{D} = \underbrace{\mathbb{E}_{\Tilde{x}\sim\mathbb{P}_g}[D(\Tilde{x})]-\mathbb{E}_{x\sim\mathbb{P}_{r}}[D(x)]}_{\text{Wasserstein loss}}+\underbrace{\tau\mathbb{E}_{\hat{x}\sim\mathbb{P}_{\hat{x}}}[(||\nabla_{\hat{x}}D(\hat{x})||_{2}-1)^{2}]}_{\text{Gradient penalty}}
\end{equation}

\begin{equation}
    \mathcal{L}_{G}= -\mathbb{E}_{\Tilde{x}\sim\mathbb{P}_g}[D(G(\Tilde{x}))]
\end{equation}

where $D$ is the set of 1-Lipschitz functions defining the discriminator network, $G$ represents the generator network and $\tau$ is the penalty coefficient.\\
\\
\textbf{Classification $\&$ Information Losses\cite{tablegan}}-\label{Ch4:tabloss} The \textbf{classification loss} requires to add to the GAN architecture an \textit{auxiliary classifier} in parallel to the discriminator. Moreover, the \textit{auxiliary classifier} is trained alongside the discriminator and generator and usually features the same neural architecture as the discriminator~\cite{acgan}. It's primarily used to output predicted class labels for each synthesized record. 

The \textit{classification loss} quantifies the discrepancy between the synthesized and predicted class labels. This helps to increase the semantic integrity of synthetic records. For instance, (sex=female, disease=prostate cancer) is not a semantically correct record as women do not have a prostate, and no such record should appear in the original data and is hence not learnt by the classifier\cite{tablegan}. Therefore, it provides the generator a useful signal to generate valid class labels for synthetic data records. 

$\mathcal{L}_{class}^{C} = \mathbb{E}[|l(x)-\mathcal{C}(fe(x))|]_{x \sim p_{data}(x)}$
$\&$ $\mathcal{L}_{class}^{G} = \mathbb{E}[|l(G(z))-\mathcal{C}(fe(G(z)))|]_{z \sim p(z)}$ correspond to training the classifier (i.e., $\mathcal{C}$) and generator (i.e., $\mathcal{G}$), respectively, where $l(.)$ is a function that returns the class label of any given data row and $fe(.)$ deletes the class feature of that data row.\\
\\
The \textbf{information loss} penalizes the discrepancy between statistics of the generated data and the real data. This helps to generate data which is statistically closer to the real one. Moreover, the information loss stabilizes the training of the generator by providing a new objective for the generator that prevents it from over-training on the current discriminator~\cite{salimans2016improved}. 

For computing the \textit{information loss}, let $f_x$ and $f_{\mathcal{G}(z)}$ denote the resulting features obtained from the penultimate layer of a discriminator denoted as $\mathcal{D}$ for a real and generated sample, respectively.  Thus, the \textit{information loss} for the generator (i.e., $\mathcal{G}$) is expressed as:
$\mathcal{L}_{info}^{G}= \mathcal{L}_{mean} + \mathcal{L}_{sd}$ where $\mathcal{L}_{mean} = ||\mathbb{E}[f_x]_{x \sim p_{data}(x)} - \mathbb{E}[f_{\mathcal{G}(z)}]_{z \sim p(z)}||_{2}$ and $\mathcal{L}_{sd} = ||\mathbb{SD}[f_x]_{x \sim p_{data}(x)} - \mathbb{SD}[f_{\mathcal{G}(z)}]_{z \sim p(z)}||_{2}$. And, $\mathbb{E}$ and $\mathbb{SD}$ denote the mean and standard deviations of the features, respectively.

Note that for all the above loss equations, $p_{data}$ is used to denote the real data distribution and $p(z)$ is a prior distribution over the latent noise vector $z$ that is fed to the generator. 

\subsection{Data Transformation}

\textbf{Mode-Specific Normalisation}-\label{Ch4:msn} The \textbf{mode-specific normalization (MSN)}~\cite{ctgan} technique developed in the work of \cite{ctgan} is invented to deal with multiple peaks in \textit{multi-modal} continuous variables. The MSN acts as a \textit{reversible transformation} that helps to represent complicated numerical distributions and generate synthetic data with greater fidelity. 
\begin{figure}[htb]
	\begin{center}
		\subfloat[\centering Fitting VGM on a continuous variable ]{
			\includegraphics[width=0.20\columnwidth]{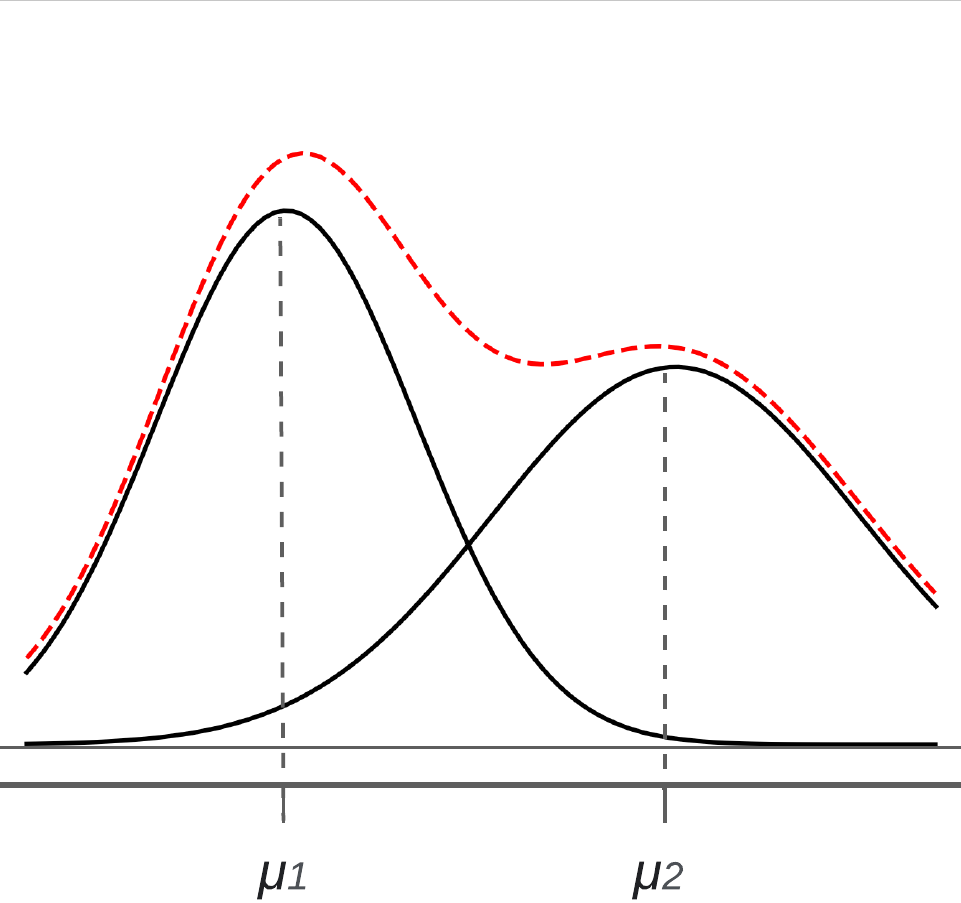}
			\label{fig:vgm}
		}
		\subfloat[\centering Selecting a mode for a single value in a continuous variable]{
			\includegraphics[width=0.20\columnwidth]{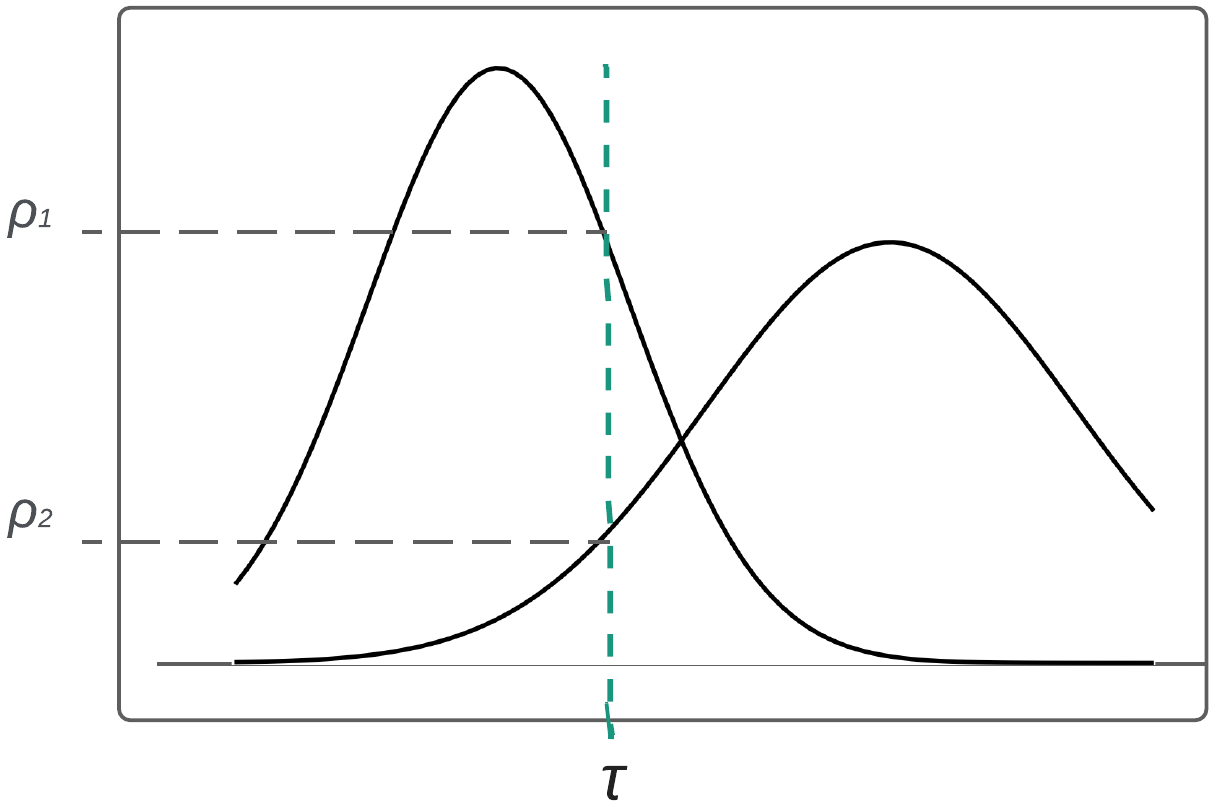}
			\label{fig:vgm_single}
		}
		\caption{\centering MSN encoding for continuous variables}
		\label{fig:gmm_distribution_continuous}
	\end{center}
\end{figure}

A continuous variable is processed using a \textit{variational Gaussian mixture model (VGM)}~\cite{prml} to estimate the number of modes $k$, e.g., $k=2$ in the example provided(see Fig.~\ref{fig:gmm_distribution_continuous}(a)), and fits a Gaussian mixture model. The learned Gaussian mixture model can be formally expressed as: $\mathbb{P} = \sum_{k=1}^{2} \omega_k \mathcal{N}(\mu_k, \sigma_k)$, where $\mathcal{N}$ is the normal distribution and $\omega_k$, $\mu_k$ and $\sigma_k$ are the weight, mean and standard deviation of each mode, respectively.

To encode values of a continuous variable, each value is associated and normalized based on the mode for which it has highest probability to belong to (see Fig.~\ref{fig:gmm_distribution_continuous}(b)). Given $\rho_1$ and $\rho_2$ being the probability density from the two modes in correspondence of the value $\tau$ to encode, the mode with the highest probability is selected. In the provided example $\rho_1$ is higher and so mode $1$ is used to normalize $\tau$. The normalized value $\alpha$ is: $\alpha =  \frac{\tau - \mu_1}{4\sigma_1}$. Moreover the mode used to encode $\tau$ is tracked via one-hot encoding $\beta$, e.g.  $\beta = [1,0]$ in the given example. The final encoding is giving by the concatenation of $\alpha$ and $\beta$: $\alpha \bigoplus \beta$, where $\bigoplus$ is the vector concatenation operator.

\subsection{Differential Privacy}
\label{Ch5:background}
This section presents formal definitions and theorems pertaining to differential privacy that are relevant for this work.\\
\\
\textbf{Definition 2.2.1} (Differential Privacy\cite{dwork2008differential}) A randomized mechanism \( \mathcal{M} \) with range \( \mathcal{R} \) is $(\epsilon,\delta)$-DP, if
\begin{equation}
    P[\mathcal{M}(S)\in \mathcal{O}] \leq e^{\epsilon}.P[\mathcal{M}(S')\in \mathcal{O}] + \delta
\end{equation}
holds for any subset of outputs  $\mathcal{O} \subseteq \mathcal{R}$ and for any adjacent datasets S and S', where S and S' differ from each other with only one training example. 

Note that, $\mathcal{M}$, for the purposes of this work corresponds to a tabular GAN model and $(\epsilon,\delta)$ represents the privacy budget. Intuitively, DP tries to minimize the influence of any individual data point on the training of tabular GANs with lower values of $(\epsilon,\delta)$ providing greater privacy protection. 
\\
\\
\textbf{Definition 2.2.2} (R\'enyi Differential Privacy (RDP)\cite{mironov2017renyi}) A randomized mechanism \( \mathcal{M} \) is $(\lambda,\epsilon)$-RDP with order $\lambda$, if 
\begin{equation}
D_{\lambda}(\mathcal{M}(S)||\mathcal{M}(S')) = \frac{1}{\lambda-1}log\mathbb{E}_{x\sim\mathcal{M}(S)} \left[  \left(\frac{P[\mathcal{M}(S)=x]}{P[\mathcal{M}(S')=x]} \right) \right]^{\lambda-1}\leq\epsilon
\end{equation}
holds for any adjacent datasets S and S', where
\small
$D_\lambda(P||Q)=\frac{1}{\lambda-1}log\mathbb{E}_{x\sim Q}[(P(x)/Q(x))^\lambda]$ \normalsize represents the R\'enyi divergence. In addition, a $(\lambda,\epsilon)$-RDP mechanism \( \mathcal{M} \) can be expressed as $(\epsilon+\frac{log1/\delta}{\lambda-1},\delta)$\normalsize-DP. 

RDP was proposed to alleviate the shortcomings of DP while dealing with the composition of randomized mechanisms that rely on the application of gaussian noise. RDP is a strictly stronger privacy definition than DP as it provides tighter bounds for tracking the cumulative privacy loss over a sequence of mechanisms such as differential private stochastic gradient descent which is performed multiple times during training.
\\
\\
\textbf{Theorem 2.2.1} (Composition\cite{mironov2017renyi}) For a sequence of mechanisms $ \mathcal{M}_{1},...,\mathcal{M}_{k}$ such that $\mathcal{M}_{i}$ is $(\lambda,\epsilon_i)$-RDP $\forall i$, the composition  $\mathcal{M}_{1}\circ ... \circ \mathcal{M}_{k}$ is $(\lambda,\sum_{i}\epsilon_{i})$-RDP.
\\
\\ 
\textbf{Definition 2.2.3} (Gaussian Mechanism\cite{dwork2014algorithmic,mironov2017renyi}) Let $f : X \rightarrow \mathbb{R}^{d}$ be an arbitrary d-dimensional function with sensitivity being:
\begin{equation}
    \Delta_{2}f = \max_{S,S'}||f(S) - f(S')||_{2}
\end{equation}
over all adjacent datasets S and S'. The Gaussian Mechanism \( \mathcal{M}_{\sigma} \), parameterized by $\sigma$, adds into the output, i.e.,
\begin{equation}
    \mathcal{M}_{\sigma}(x) = f(x) + \mathcal{N}(0,\sigma^{2}I)
\end{equation}
where $\mathcal{N}$ denotes a Gaussian distribution with mean 0 and covariance $\sigma^{2}I$. Thus, \( \mathcal{M} \) is considered to be $(\lambda,\frac{\lambda\Delta_{2}f^{2}}{2\sigma^{2}})$-RDP. 

The Gaussian mechanism described above forms the basis on which differential privacy is integrated for training tabular GANs in this work. 
\\
\\
\textbf{Theorem 2.2.2} (Post Processing\cite{dwork2014algorithmic}) If \( \mathcal{M} \) satisfies $(\epsilon,\delta)$-DP, $F\circ\mathcal{M}$ will satisfy $(\epsilon,\delta)$-DP
for any function F with $\circ$ denoting the composition operator. 

As a result of the post processing theorem, it suffices to ensure that one of the networks for tabular GANs (i.e., either the discriminator or the generator network) is trained with DP guarantees to guarantee that the overall algorithm is compatible with differential privacy.
\\
\\
\textbf{Theorem 2.2.3} (RDP for Subsampled Mechanisms\cite{wang2019subsampled}) Given a dataset containing $n$ data points with domain $\mathcal{X}$ and a randomized mechanism $\mathcal{M}$ that takes an input from $\mathcal{X}^{m}$ for $m \geq n$, let the randomized algorithm $\mathcal{M}\circ\textbf{subsample}$ be defined as: (i) \textbf{subsample}: subsample without replacement $m$ data points of the database (with subsampling rate $\gamma = m/n$); (ii) apply $\mathcal{M}$: a randomized algorithm taking the subsampled dataset as the input. 
Thus, for all integers $\lambda \geq 2 $, if $\mathcal{M}$ is $(\lambda,\epsilon(\lambda))$-RDP, then $\mathcal{M}\circ\textbf{subsample}$ is $(\lambda,\epsilon'(\lambda))$-RDP where

\begin{equation}
\begin{aligned}
& \epsilon'(\lambda) & \leq \frac{1}{\lambda-1}log\bigg(1 + \gamma^{2}\binom{\lambda}{2}\min\left\{4(e^{\epsilon(2)}-1),e^{\epsilon(2)}\min\{2,(e^{\epsilon(\infty)}-1)^{2}\}\right\} \\
&  & + \sum_{j=3}^{\lambda}\gamma^{j} \binom{\lambda}{j}e^{(j-1)\epsilon(j)}\min\{2,(e^{\epsilon(\infty)-1)^{j})}\}\bigg)
\end{aligned}
\end{equation}

Subsampling is a useful technique to strengthen the privacy guarantees offered by a randomized mechanism $\mathcal{M}$. 

\subsection{DP via Differential Private SGD\cite{abadi2016deep}}
\label{Ch5:DPSGD}
The DP-SGD technique enables training neural networks with differential privacy guarantees and uses noisy stochastic gradient descent as a means to limit the influence of individual training samples. Algorithm 1 specifies how this technique is used for training a network with parameters $\theta$ by minimizing the empirical loss function $\mathcal{L}(\theta)$. For every iteration of SGD, the gradients $\Delta\mathcal{L}(
\theta; x_i)$ are calculated for some random subset of real data points. After which the L2 norm of the gradients are clipped. Finally noise is added to the gradients to preserve privacy and the parameters $\theta$ are updated via gradient descent. 

\begin{algorithm}[htb]
\textbf{Input:} Data points $\{x_1,...,x_N\}$, loss function \small$\mathbb{\mathcal{L}}(\theta) = \frac{1}{N}\sum_{i}\mathbb{\mathcal{L}}(\theta,x_i)$. \normalsize 
 Hyper-parameters: learning rate $\eta_{t}$, noise scale $\sigma$, batch size $B$, gradient norm bound $C$.\\
\textbf{Initialize} $\theta_{0}$ randomly\;
 
\For{$t \in [T]$}{
  Take a random sample $B_t$ with sampling probability $B/N$\;
    
  \textbf{Compute Gradient}\:
  
  For each $i \in B_{t}$, compute $g_{t}(x_{i})\leftarrow\nabla_{\theta_{t}}\mathcal{L}(\theta_{t},x_{i})$\:
  
  \textbf{Clip gradient}\:
  
  $\bar{g}_{t}\leftarrow g_{t}(x_i)/\max(1,\frac{||g_t(x_i)||_{2}}{C})$
  
  \textbf{Add noise}\:
  
  $\Tilde{g}_{t}\leftarrow\frac{1}{L}(\sum_{i}\bar{g}_{t}(x_i)+\mathcal{N}(0,\sigma^{2}C^{2}I))$
  
  \textbf{Descent}\:
  
  $\theta_{t+1}\leftarrow\theta_t-\eta_t\Tilde{g}_t$
  
 }
\textbf{Output:}$\theta_{T}$ and final privacy cost $(\epsilon,\delta)$ computed using a privacy accountant.

\caption{\centering Differential Private SGD\cite{abadi2016deep}}
\label{Ch5:DPSGD-Algo}
\end{algorithm}

%% file: Content/Chapters/3_Exploratory_Study.tex
\chapter{Exploratory Study of Related Work}\label{ch3}

\section{Introduction}

This chapter tackles the first research question introduced in Sec.~\ref{Ch1:research_question}. Thus, Sec.~\ref{Ch3:EC} elicits an empirical quantitative analysis of four state-of-the-art tabular GANs with respect to three core evaluation criteria as follows:

\begin{itemize}
    \item Statistical similarity with original data 
    \item Utility for Machine Learning (ML) applications
    \item Privacy preservability
\end{itemize}
And, Sec.~\ref{Ch3:Challenges} introduces major challenges faced by current state-of-the-art methods. Finally, Sec.~\ref{Ch3:Conclusion} ends the chapter with a brief summary. 

\section{Empirical Comparison}
\label{Ch3:EC}

\subsection{Datasets}
\label{Ch3:DD}

Five commonly used machine learning datasets were used to perform this experimental study,. Three of them -- \href{http://archive.ics.uci.edu/ml/datasets/adult}{Adult}, \href{https://archive.ics.uci.edu/ml/datasets/covertype}{Covertype} and \href{http://archive.ics.uci.edu/ml/datasets/kdd+cup+1999+data}{Intrusion} -- are from the UCI machine learning repository~\cite{UCIdataset}. The other two --\href{https://www.kaggle.com/mlg-ulb/creditcardfraud}{Credit} and \href{https://www.kaggle.com/itsmesunil/bank-loan-modelling}{Loan} -- are from Kaggle\footnote{https://www.kaggle.com/datasets}. All five tabular datasets have a target variable, for which the rest of the variables are used to perform classification. Due to computing resource limitations, 50K rows of data are sampled randomly in a stratified manner with respect to the target variable for Covertype, Credit and Intrusion datasets.

However, the Adult and Loan datasets are not sampled. The details of each dataset are shown in Tab.~\ref{table:DDE}.  One thing to notice here is that we assume that the user already knows the data type of each variable for every dataset before training. \cite{ctgan} holds the same assumptions.

\begin{table}[htb]
\centering
\caption[DD]{\centering Description of Datasets\footnotemark.}
\resizebox{0.8\columnwidth}{!}{
\begin{tabular}{ |c|c|c|c|c|c|c|c| }
\hline
\textbf{Dataset} & \textbf{Train/Test Split} &\textbf{Target Variable} & \textbf{$\mbox{Continuous}$}  & \textbf{$\mbox{Binary}$} & \textbf{$\mbox{Multi-class}$} & \textbf{$\mbox{Mixed-type}$}& \textbf{$\mbox{Long-tail}$}\\ 
\hline
{Adult}     & 39k/9k   & 'income'        & 3   & 2  & 7  & 2 & 0\\
\hline
 Covertype & 45k/5k    & 'Cover\_Type'    & 10   & 44 & 1  & 0 & 0\\\hline
 Credit    & 40k/10k    & 'Class'         & 30  & 1 & 0  & 0 & 1 \\\hline
 Intrusion & 45k/5k    & 'Class'         & 22   & 6 & 14 & 0 & 2 \\\hline
 Loan      & 4k/1k       & 'PersonalLoan'  & 5   & 5  & 2  & 1 & 0 \\
\hline
\end{tabular}
}
\label{table:DDE}
\end{table}

\footnotetext{ Refer to Sec.~\ref{Ch4:data_representation} $\&$ Sec.~\ref{Ch4:longgtail} for details for Mixed-type and Long-tail, respectively. Note that these data-types are simply treated as continuous with respect to the baselines evaluated in this chapter.
}
\subsection{Baselines}
\label{Ch3:Baselines}

We evaluate 4 state-of-the-art GAN-based tabular data generators: CTGAN, TableGAN, CW-GAN $\&$ MedGAN. Tab.~\ref{table:sota} stresses on the key features of each baseline. 

To have a fair comparison, all algorithms are implemented in Pytorch, with the generator and discriminator structures matching the descriptions provided in their respective papers with the exception of the MedGAN model which was extended to deal with continuous variables as well.\footnote{Note that the code-base for all the models was found here- \url{https://github.com/sdv-dev/SDGym}} All algorithms are trained using a batch size of 500 rows for 150 epochs for Adult, Covertype, Credit and Intrusion datasets, whereas the algorithms are trained for 300 epochs on Loan dataset. This is because, the Loan dataset is significantly smaller than the others containing only 5000 rows and requires a longer training time to converge. Lastly, each experiment is repeated 3 times.

\subsection{Environment}

Experiments are run under Ubuntu 20.04 on a machine equipped with 32 GB memory, a GeForce RTX 2080 Ti GPU and a 10-core Intel i9 CPU.

\subsection{Evaluation metrics}
\label{Ch3:metrics}

The evaluation is conducted on three dimensions: (1) machine learning (ML) utility, (2) statistical similarity and (3) privacy preservability. The first two are used to evaluate if the synthetic data can be used as a good proxy of the original data. The third criterion sheds light on the nearest neighbour distances within and between the original and synthetic datasets, respectively.
 
 \begin{enumerate}
     \item \textbf{Machine learning (ML) utility}-
\label{Ch3:ml_efficacy}
To quantify the ML utility, we compare the performance achieved by 5 widely used machine learning algorithms on real versus synthetic data: decision tree classifier, linear support-vector-machine (SVM), random forest classifier, multinomial logistic regression and MLP. We use Python and scikit-learn 0.24.2. 
We set max-depth to 28 for decision tree and random forest models. MLP uses one 128 neuron hidden layer. All other hyper-parameters use their default value. For a fair compassion, all hyper-parameters and ML models are fixed across all datasets. Due to this our results can differ slightly from~\cite{ctgan} where the authors use different ML models and hyper-parameters for each dataset.

First we split the original data into training and test sets (see Fig.~\ref{fig:settingA}). The training set is used as real data to train the GAN models. Once the training is finished, we use it to synthesize data with the same size as the training set. The synthetic and real training datasets are then used to train two separate instances of the 5 machine learning models from above. The ML utility is measured via difference in accuracy, F1-score and area under the ROC between model pairs trained on the real and synthetic data. The aim of this design is to test how close the ML utility is when we train a machine learning model using the synthetic data vs the real data. 

\begin{figure}[htb]
    \centering
    \includegraphics[width=0.7\columnwidth]{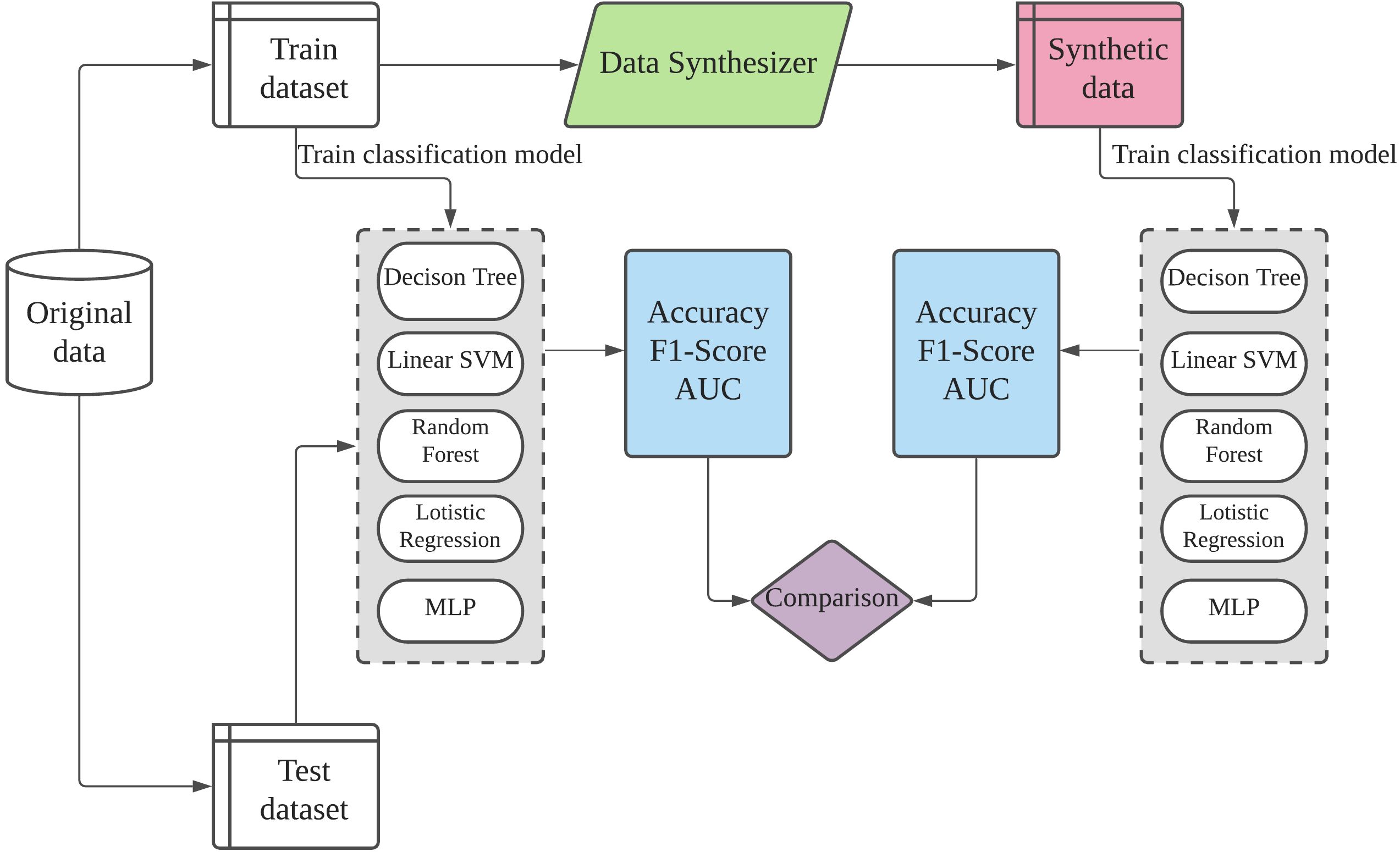}
    \caption{\centering Evaluation flows for ML utility}
    \label{fig:settingA}
    \vspace{-0.5em}
\end{figure}  

\item \textbf{Statistical Similarity}- Three metrics are used to quantitatively measure the statistical similarity between the real and synthetic data.

\textit{Jensen-Shannon divergence (JSD)}~\cite{jsd}- The JSD provides a measure to quantify the difference between the probability mass distributions of individual categorical variables belonging to the real and synthetic datasets, respectively. Moreover, this metric is bounded between 0 and 1 and is symmetric allowing for an easy interpretation of results.

\textit{Wasserstein distance (WD)}~\cite{wgan_test}- In similar vein, the Wasserstein distance is used to capture how well the distributions of individual continuous/mixed variables are emulated by synthetically produced datasets in correspondence to real datasets. We use WD because we found that the JSD metric was numerically unstable for evaluating the quality of continuous variables, especially when there is no overlap between the synthetic and original dataset. Hence, we resorted to utilize the more stable Wasserstein distance. 
 
\textit{Difference in pair-wise correlation (Diff. Corr.)}- 
To evaluate how well feature interactions are preserved in the synthetic datasets, we first compute the pair-wise correlation matrix for the columns within real and synthetic datasets individually. To measure the correlation between any two continuous features, the Pearson correlation coefficient is used. It ranges between $[-1,+1]$. Similarly, the Theil uncertainty coefficient is used to measure the correlation between any two categorical features. It ranges between $[0,1]$. Lastly, the correlation ratio  between categorical and continuous variables is used. It also ranges between $[0,1]$. Note that the dython\footnote{\url{http://shakedzy.xyz/dython/modules/nominal/\#compute\_associations}} library is used to compute these metrics. Finally, the differences between the pair-wise correlation matrices for the real and synthetic datasets is computed.

\item \textbf{Privacy preservability}- To quantify the privacy preservability, we resort to distance metrics (instead of differential privacy~\cite{pategan}) as they are intuitive and easy to understand by data science practitioners. Specifically, the following two metrics are used to evaluate the privacy risk associated with synthetic datasets.

\textit{Distance to Closest Record (DCR)}- The DCR is used to measure the Euclidean distance between any synthetic record and its closest corresponding real neighbour. Ideally, the higher the DCR the lesser the risk of privacy breach. Furthermore, the $5^{th}$ percentile of this metric is computed to provide a robust estimate of the privacy risk.
    
\textit{Nearest Neighbour Distance Ratio (NNDR)}~\cite{nndr}- Instead of only measuring the closest neighbour, the NNDR measures the ratio between the Euclidean distance for the closest and second closest real neighbour to any corresponding synthetic record. This ratio is within $[0,1]$. Higher values indicate better privacy. Low NNDR values between synthetic and real data may reveal sensitive information from the closest real data record. Fig.~\ref{fig:nndr} illustrates the case.  Hence, this ratio helps to evaluate the privacy risk with greater depth and better certainty. Note that the $5^{th}$ percentile is computed here as well.

\begin{figure}[htb]
    \centering
    \includegraphics[width=0.5\columnwidth]{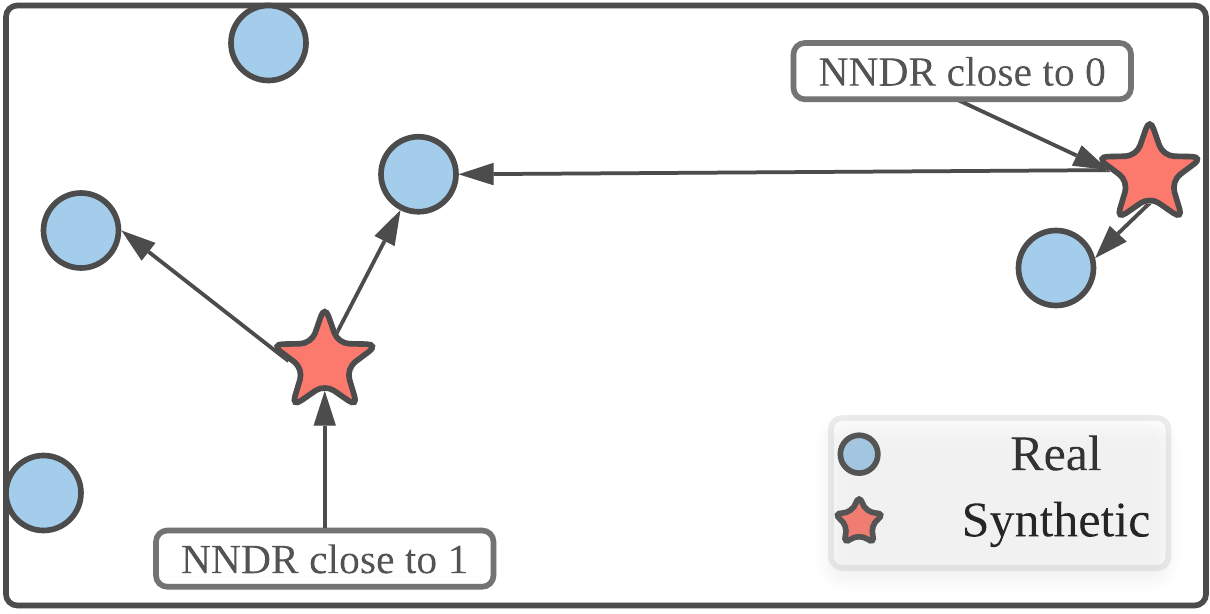}
    \caption{ \centering Illustration of NNDR metric with its privacy risk implications}
    \label{fig:nndr}
\end{figure}

 \end{enumerate}

\subsection{Results}
\label{Ch3:Results}
 
 In this sub-section, the experimental results for each data-synthesizer are shown based on the aforementioned evaluation criteria.
 
 \begin{enumerate}
     \item \textbf{ML Utility}- Tab. \ref{table:ML_allE} shows that the TableGAN model outperforms the other models by achieving the least differences in all three metrics used to measure ML utility (i.e Accuracy, F1-score and AUC). This surprising result shows that it can even outperform the more recent conditional GAN architectures such as CTGAN and CW-GAN. The results shown here suggests that the deep convolution architecture employed in the TableGAN model achieves the most formidable results. Therefore, it is worth exploring the benefits of utilising this type of architecture to further improve the performance of other models such as CTGAN.

\begin{table}[htb]
\centering
\caption{\centering Difference of ML accuracy (\%), F1-score, and AUC between original and synthetic data: average over 5 different datasets and 3 replications.}
\begin{tabular}{|c|c|c|c|}
\hline
\textbf{Method} & \textbf{Accuracy} & \textbf{F1-score}  & \textbf{AUC} \\
\hline
\small{CTGAN}    &21.51\% & 0.274 &   0.253 \\
\small{TableGAN} &\textbf{11.40\%}  & \textbf{0.130} &  \textbf{0.169} \\
\small{MedGAN}    & 14.11\%&  0.282&   0.285 \\
\small{CW-GAN}    &  20.06\%& 0.354 &  0.299 \\
\hline
\end{tabular}
\label{table:ML_allE}
\end{table}

\item \textbf{Statistical similarity}- Tab. \ref{table:SS_allE} shows that the CTGAN data-synthesizer achieves the best average JSD for categorical columns along with the the best average wasserstein distance for continuous columns. This highlights that a conditional architecture accompanied by the training-by-sampling method along with mode specific normalisation for continuous variables is directly beneficial for improving statistical similarity of synthetically produced datasets. However, it should be noted the TableGAN model performs best in terms of maintaining the least correlation distance with the CTGAN model performing slightly worse at second place. This slight difference could yet again be attributed towards the \textit{DCGAN neural network architecture} that makes use of \textit{strided convolutions} which enables the receptive field to grow larger after each layer thereby extracting useful global correlations in the data.

\begin{table}[htb]
\centering
\caption{\centering Statistical similarity: three measures averaged over 5 datasets and three repetitions.}
\begin{tabular}{|c|c|c|c|}
\hline
\textbf{Method} & \textbf{Avg JSD} & \textbf{Avg WD}  & \textbf{Diff. Corr.} \\
\hline
\small{CTGAN}    & \textbf{0.0704} &  \textbf{1769}  & 2.73  \\
\small{TableGAN} & 0.0796& 2117   &  \textbf{2.30} \\
\small{MedGAN}   & 0.2135& 46257  &   5.48   \\
\small{CW-GAN}   & 0.1318& 238155 &  5.82 \\
\hline
\end{tabular}
\label{table:SS_allE}
\vspace{-0.5em}
\end{table}

\item \textbf{Privacy Impact}- Tab. \ref{table:PP_allCh3} highlights that the CW-GAN and MedGAN models maintain the safest distance in terms of the DCR and NNDR metrics between real and synthetic datasets. Furthermore, by analyzing the DCR and NNDR metrics within synthetic data, we see that the CW-GAN model produces the most diverse samples whereas the MedGAN model produces the least diverse samples among all the data-synthesizers suggesting that it most likely suffers from mode-collapse. Lastly, it is worth mentioning that the results also show that privacy and ML utility are fundamentally inversely related as models such as CTGAN and TableGAN which perform well in terms of ML utility are naturally worse in terms of privacy.  

\begin{table}[htb]
\centering
\caption{\centering Privacy impact: between real and synthetic data (R\&S) and within real data (R) and synthetic data (S).}
\begin{tabular}{|c|c|c|c|c|c|c|}
\hline
\multirow{2}{*}{\textbf{Model}} & \multicolumn{3}{c|}{\textbf{DCR}} & \multicolumn{3}{c|}{\textbf{NNDR}} \\
\cline{2-7}
 & \textbf{R\&S} & \textbf{R}  & \textbf{S} & \textbf{R\&S} & \textbf{R}  & \textbf{S}\\

\hline
CTGAN    & 1.517 &  0.428& 1.026 & 0.763 & 0.414 &0.624 \\
TableGAN & 0.988 &  0.428& 0.920 & 0.681 & 0.414 &0.632\\
MedGAN   & 1.918 &  0.428& 0.254 & \textbf{0.871} & 0.414 &0.393 \\
CW-GAN   & \textbf{2.197} &  0.428& 1.124 & 0.847 & 0.414 &0.675\\
\hline
\end{tabular}
\label{table:PP_allCh3}
\end{table}

\end{enumerate}

\section{Challenges faced by Existing Solutions}
\label{Ch3:Challenges}

In this section, we empirically demonstrate how the prior state-of-the-art methods fall short in solving challenges in industrial data sets. 

\begin{figure}[htb]
	\begin{center}
		\subfloat[\centering Mortgage in Loan dataset~\cite{kaggleloan}]{
			\includegraphics[width=0.33\textwidth]{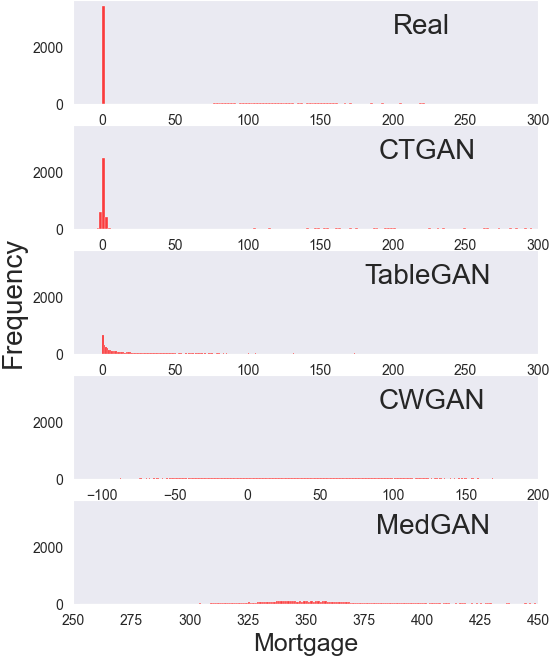}
			\label{fig:mortgage_column_motivation}
		}
		\subfloat[\centering Amount in Credit dataset\cite{kagglecredit}]{
			\includegraphics[width=0.3\textwidth]{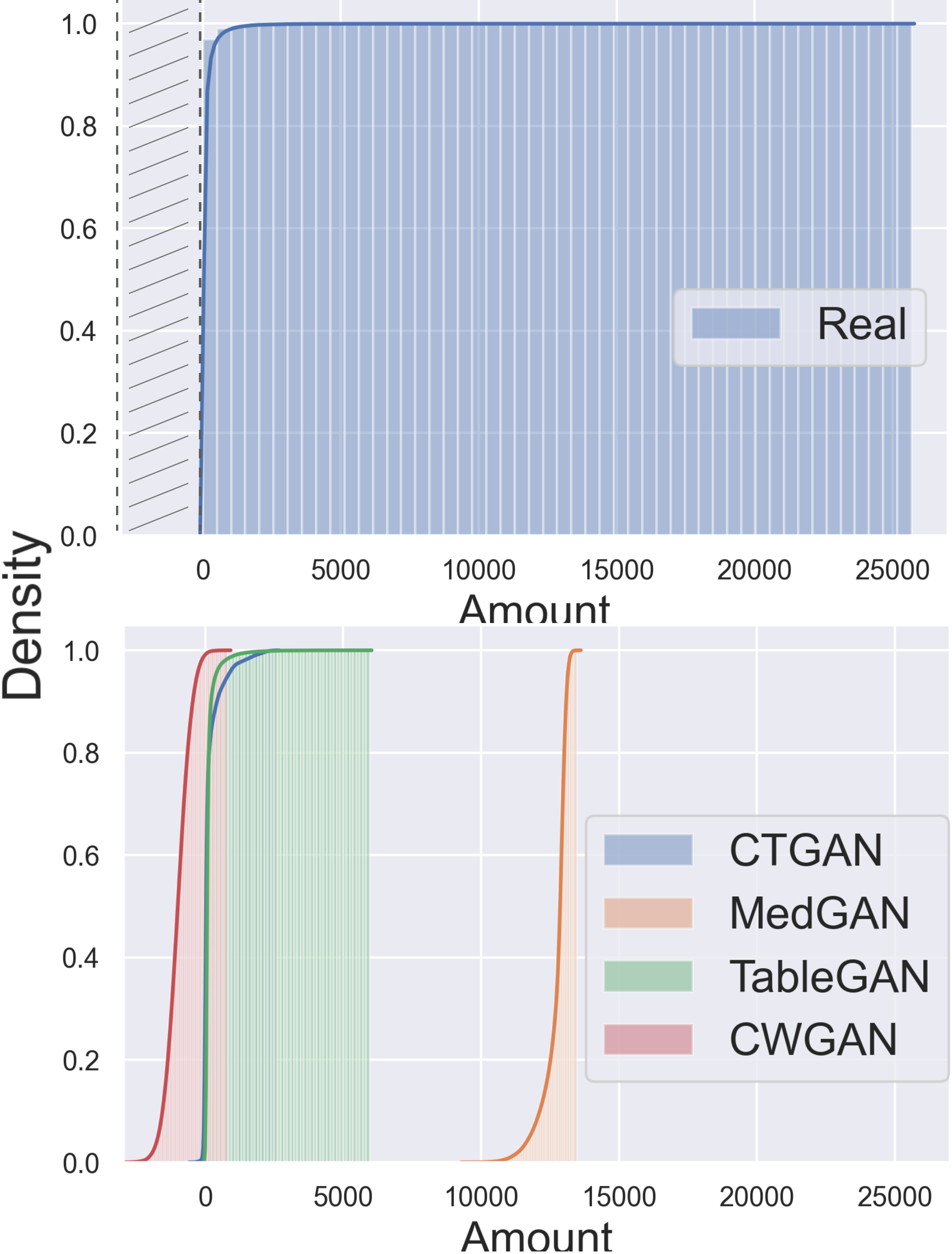}
			\label{fig:amount_result_motivation}
		}
		\subfloat[\centering Hours-per-week in Adult dataset~\cite{UCIdataset}]{
			\includegraphics[width=0.335\textwidth]{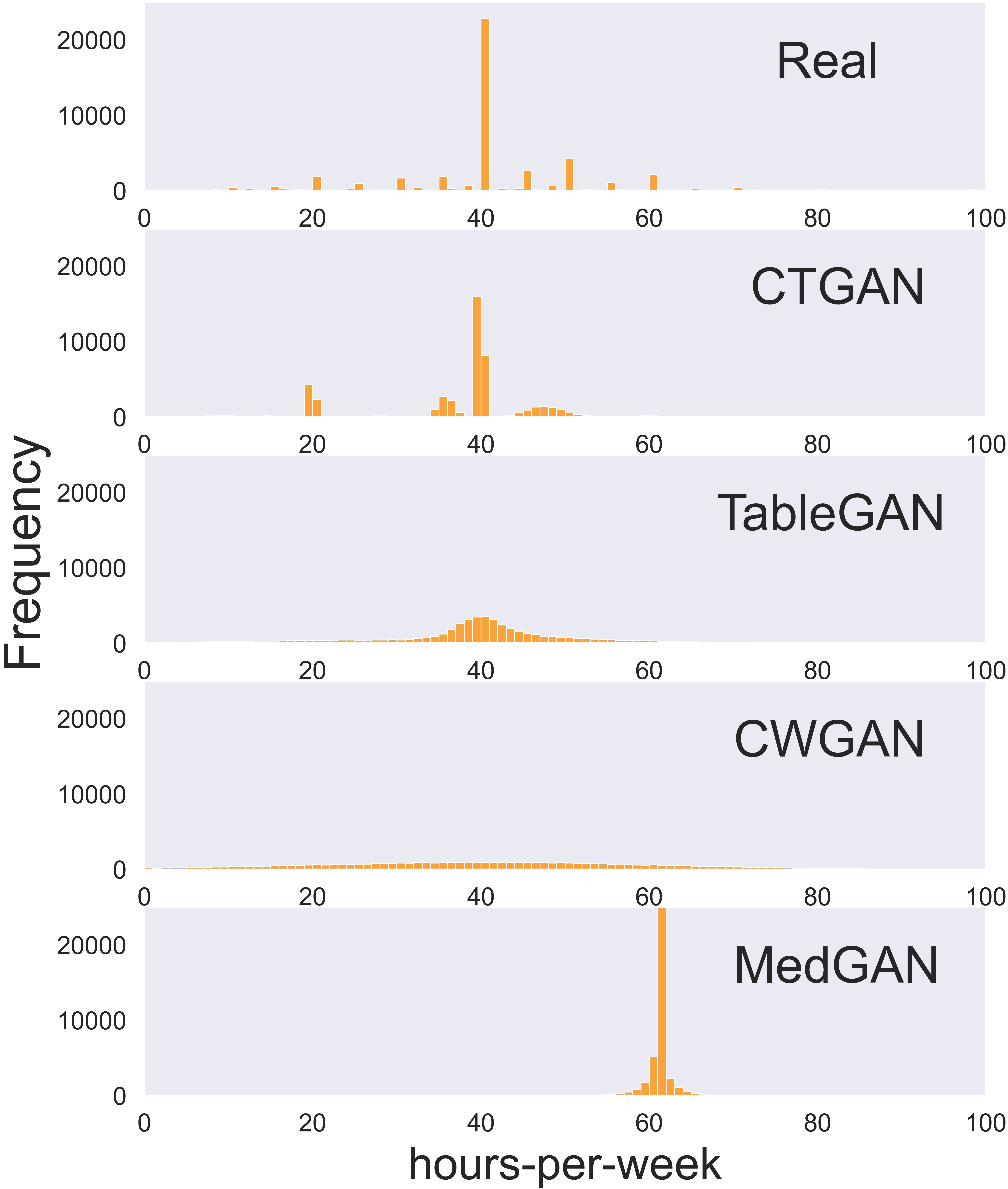}
			\label{fig:gmm_result_motivation}
		}
		\caption{\centering Challenges of modeling industrial datasets using existing Tabular  GANs: (a) Mixed data-type, (b) Long tail distribution, and (c) Skewed multi-modal continuous variable} 
		\label{fig:motivationcases}
 	\end{center}
\end{figure}

\begin{enumerate}
    \item \textbf{Mixed data type variables}- To the best of our knowledge, existing GAN-based tabular generators only consider data variables as either categorical or continuous. However, in reality, a variable can be a mix of these two types, and often variables have missing values. The \textit{Mortgage} variable from the Loan dataset is a good example of a mixed variable. Fig.~\ref{fig:mortgage_column_motivation} shows the distribution of the original and synthetic data generated by 4 state-of-the-art algorithms for this variable.

    According to the data description, a loan holder can either have no mortgage (0 value) or a mortgage (any positive value). In appearance this variable is not a categorical type due to the numeric nature of the data. So all 4 state-of-the-art algorithms treat this variables as a continuous type without capturing the special meaning of the value, zero. Hence, all 4 algorithms generate a value around 0 instead of exact 0. And the negative values for Mortgage have no/wrong meaning in the real world. 
    
    \item \textbf{Long tail distributions}- Many real world datasets can have long tail distributions where most of the occurrences happen near the initial value of the distribution, and rare cases towards the end. As an example, Fig.~\ref{fig:amount_result_motivation} plots the cumulative frequency for the original (top) and synthetic (bottom) data generated by 4 state-of-the-art algorithms for the \textit{Amount} variable in the Credit dataset. This variable represents the transaction amount when using credit cards. One can imagine that most transactions have small amounts, ranging from few bucks to thousands of dollars. However, there definitely exists a very small number of transactions with large amounts. Note that for ease of comparison both plots use the same x-axis, but the real data has no negative values. 
    
    Thus, the real data clearly has 99\% of occurrences happening at the start of the range, but the distribution extends until around $25000$. In comparison none of the synthetic data generators are able to learn and imitate this behavior.

    \item \textbf{Skewed multi-modal continuous variables}-The term \textit{multi-mode} is extended from Variational Gaussian Mixtures (VGM)~\cite{prml} (refer to Sec.~\ref{Ch4:msn}). These are used to model Gaussian distributions with multiple peaks. The intuition behind using multiple modes can be easily captured from Fig.~\ref{fig:gmm_result_motivation}. The figure plots in each row the distribution of the working \textit{Hours-per-week} variable from the Adult dataset. This is not a typical Gaussian distribution. There is an obvious peak at 40 hours but with several other lower peaks, e.g. at 50, 20 and 45. Also the number of people working 20 hours per week is higher than those working 10 or 30 hours per week.
    
    This behavior is difficult to capture for the state-of-the-art data generators (see subsequent rows in Fig.\ref{fig:gmm_result_motivation}). The closest results are obtained by CTGAN which uses Gaussian mixture estimation for continuous variables. However, CTGAN loses some modes compared to the original distribution.
    
\end{enumerate}

\section{Conclusion}
\label{Ch3:Conclusion}

In this exploratory study, we were able to shed some light on some of the latest works on GAN-based tabular data-synthesizers. Additionally, we executed an in-depth empirical evaluation to benchmark their performance. Based on our findings, we summarize some important key points as follows-:

\begin{itemize}
\item Firstly, the TableGAN model outperforms state-of-the-art approaches with respect to the utility for ML applications as it maintains the least difference in all ML metrics. 
\item Secondly, it is observed that the best average JSD for categorical variables and the best average wasserstein distance for continuous variables is achieved by CTGAN.    
\item Thirdly, in terms of the privacy risk, all data-synthesizers produce datasets with a greater DCR and NNDR metric for between real and synthetic datasets as compared to both within real and synthetic datasets. This suggests that the privacy risk for all data-synthesizers as measured via these metrics are limited.
\item Lastly, current techniques fail to account for mixed data-types, heavy long-tailed distributions and skewed multi-modal numerical distributions. 
\end{itemize}

%% file: Content/Chapters/4_CTABGAN.tex
\chapter{CTAB-GAN: Effective Tabular Data Synthesizing}\label{ch4}

\section{Introduction}

CTAB-GAN is a novel tabular data generator designed to overcome the challenges outlined in Sec.~\ref{Ch3:Challenges}. In CTAB-GAN we invent a \textit{Mixed-type Encoder} based on the \textit{mode-specific normalization (MSN)} introduced in the work of \cite{ctgan}. The \textit{Mixed-type Encoder} can better represent a mix of categorical and continuous variables as well as deal with missing values. Moreover, CTAB-GAN is based on a \textit{conditional GAN (CGAN)} and utilizes \textit{training-by-sampling} to efficiently treat imbalanced data variables. Additionally, it features the \textit{classification}, \textit{information} and \textit{generator losses}~\cite{tablegan,ctgan} for training the generator to improve semantic integrity and training stability, respectively. Furthermore, CTAB-GAN makes use of the underlying \textit{DCGAN architecture}~\cite{radford2015unsupervised} for enhancing the quality of generated samples. Lastly, CTAB-GAN utilizes a light-weight \textit{log-transformation} to overcome the mode collapse problem for heavy long-tailed numerical variables. 

Hence, in section Sec.~\ref{Ch4:Design}, the novel design aspects of CTAB-GAN are highlighted and Sec.~\ref{Ch4:exp} provides an experimental analysis comparing CTAB-GAN with the state-of-the-art methods introduced in Sec.~\ref{Ch3:Baselines}. Lastly, Sec.~\ref{Ch4:Conclusion} ends with a short conclusion of the chapter.

\section{Design of CTAB-GAN}
\label{Ch4:Design}

\subsection{Network Structure}

\begin{figure}[htb]
    \centering
    \includegraphics[width=\linewidth]{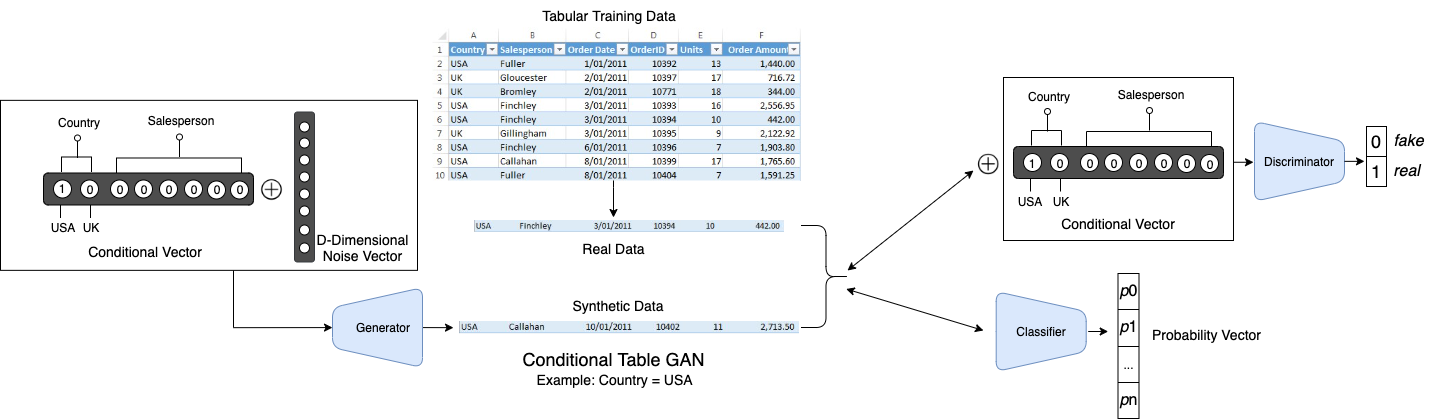}
    \caption{\centering Synthetic Tabular Data Generation via CTAB-GAN}
    \label{fig:STD_1}
\end{figure}

The structure of CTAB-GAN comprises of three blocks: Generator $\mathcal{G}$, Discriminator $\mathcal{D}$ and an auxiliary Classifier $\mathcal{C}$. Moreover, since our algorithm is based on \textit{conditional GAN}, the generator requires a noise vector plus a \textit{conditional vector} as input (refer to Sec.~\ref{Ch4:cgan}). Additionally, the discriminator is fed both the real and synthetic data after concatenating them with their corresponding \textit{conditional vectors} as input (see Fig.~\ref{fig:STD_1}).   

$\mathcal{G}$ and $\mathcal{D}$ are implemented using the \textit{DCGAN neural network architecture}~\cite{radford2015unsupervised} (refer to section \ref{Ch4:dcgan}) inspired from the work of \cite{tablegan}. This architecture has shown promising results in terms of generating synthetic data with high ML utility and was found to most optimally capture the correlations in the original data (refer to Sec.~\ref{Ch3:Results}). Therefore, it is used as the underlying neural network architecture for training CTAB-GAN. 

$\mathcal{C}$ (refer to section \ref{Ch4:tabloss}) consists of 4 fully connected layers with $256$ nodes each which are all followed by a \textit{LeakyReLU layer} with a leaky ratio of $0.20$ and are trained using \textit{dropout regularization} with a probability parameter of $0.5$. Note that the last and $5^{th}$ layer of the classifier is adapted to deal with both \textit{binary} $\&$ \textit{multi-class classification} problems. 



An important distinction concerning the classification loss as presented in the work of \cite{tablegan} is that, this work utilizes an MLP neural architecture\footnote{The MLP architecture was chosen as it led to superior performance in preliminary experiments.} for the \textit{auxiliary classifier} and caters to both \textit{binary} and \textit{multi-class classification} problems. Whereas, TableGAN features an \textit{auxiliary classifier} with the same neural architecture as the discriminator and can only deal with binary classification problems.

\subsection{Data Representation}
\label{Ch4:data_representation}

\begin{figure}[htb]
    \centering
    \includegraphics[width=0.4\linewidth]{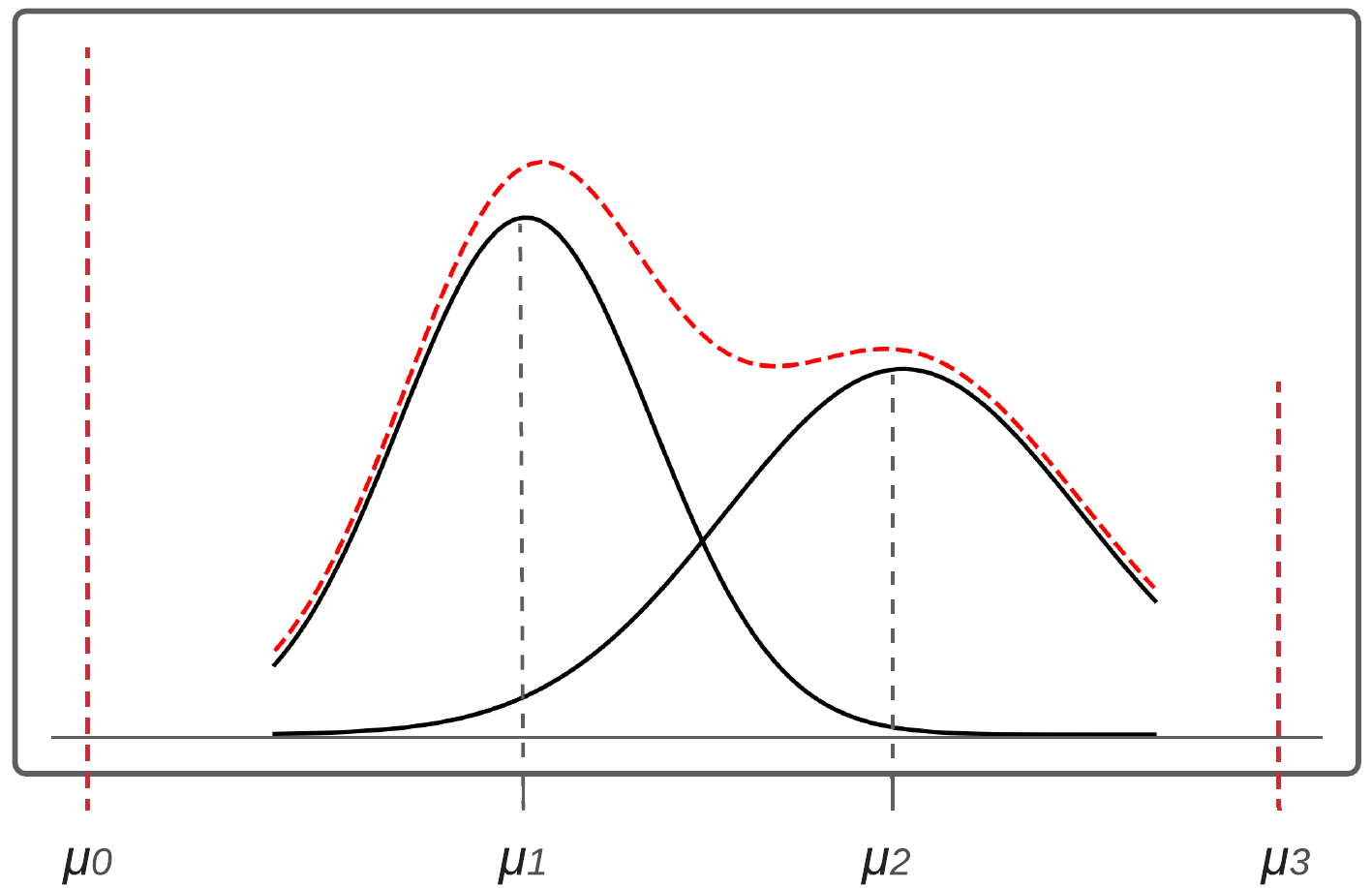}
    \caption{\centering \centering Mixed type variable distribution with VGM estimation}
    \label{fig:gmm_distribution_mixed}
\end{figure}

The original tabular training data is encoded variable by variable. This work distinguish between three types of variables: \textit{categorical}, \textit{continuous} $\&$ \textit{mixed}. 

\textbf{Mixed variables} are those that contain both \textit{categorical} and \textit{continuous} values, an example is a \textit{continuous variable} with missing values. The missing values clearly do not belong to the continuous domain. Thus, they are treated separately as a \textit{categorical component} of a \textit{mixed variable}. 

The novel \textit{mixed-type encoder} is proposed to deal with such a variable. With this encoder, values of \textit{mixed variables} are seen as concatenated value-mode pairs based on the \textit{MSN technique}(refer to Sec.~\ref{Ch4:msn}) introduced by \cite{ctgan}. The encoding is illustrated via the exemplary distribution of a \textit{mixed variable} shown in red in Fig.~\ref{fig:gmm_distribution_mixed}. One can see that values can either be exactly $\mu_0$ or $\mu_3$ (the \textit{categorical part}) or distributed around two peaks in $\mu_1$ and $\mu_2$ (the \textit{continuous part}). The \textit{continuous part} has been explained in Sec.~\ref{Ch4:msn}.

The \textit{categorical part} (e.g., $\mu_0$ or $\mu_3$) in Fig.~\ref{fig:gmm_distribution_mixed} is treated similarly, except $\alpha$ is directly set to 0 as the category is determined only by the one-hot encoding representing the modes. For example, for a value in $\mu_3$, the final encoding is given by $0 \bigoplus [0, 0, 0, 1]$.

Finally, \textit{categorical variables} are encoded via a one-hot vector $\gamma$. missing values in this case are simply treated as a separate unique class and an extra bit is added to the one-hot vector to account for it. 

Thus, a row with $[1, \dots, N]$  variables is encoded by concatenation of the encoding of all variables, i.e. either $(\alpha \bigoplus \beta)$ for \textit{continuous} $\&$ \textit{mixed variables} or $\gamma$ for \textit{categorical variables}. Having $n$ \textit{continuous/mixed variables} and $m$ \textit{categorical variables} ($n + m = N$) the final encoding can be expressed as:     
\begin{equation}
\label{condvec}
    \bigoplus_{i=1}^{n} \alpha_i\mathsmaller{\bigoplus} \beta_{i} \;
    \bigoplus_{j=n+1}^{N} \gamma_{j} 
\end{equation}

\subsection{Counter Imbalanced Variables}
\label{Ch4:condvec}

\begin{figure}[htb]
    \centering
    \includegraphics[scale=.18]{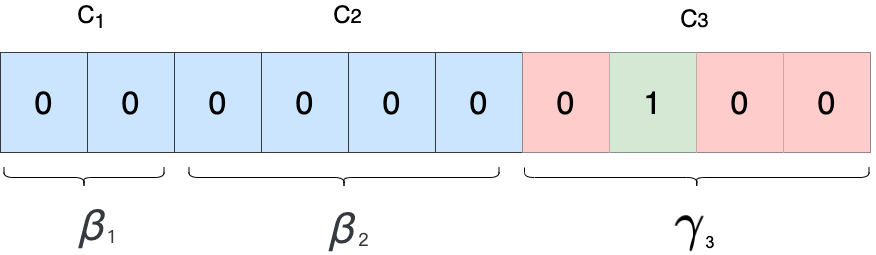}
    \caption{\centering Conditional vector: example selects class 2 from third variable out of three}
    \label{fig:condvec}
\end{figure}

In CTAB-GAN, the conditional GAN with training-by-sampling (refer to Sec.~\ref{Ch4:cgan}) inspired from the work of \cite{ctgan} is used. However, in contrast to their work, the conditional vector of CTAB-GAN is further extended to include the one-hot-vectors corresponding to the modes used to represent continuous and mixed columns (refer to Sec.~\ref{Ch4:data_representation}). Thus, the extended conditional vector $ex\_cond$ is a bit vector given by the concatenation of all one-hot encodings $\beta$ (for continuous $\&$ mixed variables) along with all categorical one-hot encodings $\gamma$ for all variables present in Eq.~\eqref{condvec}. For example, $ex\_cond$ is shown in Fig.~\ref{fig:condvec} with three  variables, one continuous ($C_1$), one mixed ($C_2$) and one categorical ($C_3$), with class 2 selected for $C_3$. 

Extending the conditional vector to include the continuous $\&$ mixed variables helps deal with imbalance in the frequency of modes used to represent them. Moreover, the generator is conditioned on all data-types during training enhancing the learned correlation between all variables (refer to Sec.~\ref{Ch4:Results} $\&$ Sec.~\ref{Ch4:motivation_response}).

\subsection{Treat Long Tails}
\label{Ch4:longgtail}

To encode continuous values, a variational Gaussian mixture model is used (as explained in Sec.~\ref{Ch4:msn}). However, Gaussian mixtures can not deal with all types of data distributions, notably distributions with a long tail where few rare points are far from the bulk of the data. VGM especially face great difficulties to encode the values towards the tail.

To counter this issue, we pre-process continuous variables with long tail distributions with a log-transform. For such a variable having values with lower bound $l$, we replace each value $\tau$ with compressed $\tau^c$:
\begin{equation}
\label{eq:preprossesing}
\tau^c =  \left\{
	\begin{array}{rl}
		 \mbox{log($\tau$)} &  \mbox{if $l>$0} \\
		\mbox{log($\tau$ - $l$+$\epsilon$)} & \mbox{if $l\leqslant$0}  \mbox{, where } \mbox{$\epsilon>$0} 	 
	\end{array} \right\}
\end{equation}

The log-transform allows to compress and reduce the distance between the tail and bulk data making it easier for VGM to encode all values, including those values present towards the end of the long tail. We show the impact of this simple yet effective method in Sec.~\ref{Ch4:motivation_response}.

\subsection{Training Procedure}
\label{Ch4:TP}

To train CTAB-GAN, one must overcome 2 major difficulties both caused by the use of a \textit{convolution} based GAN architecture (i.e., \textit{DCGAN}\cite{radford2015unsupervised}). The first is to be \textbf{data compatible} with the \textit{DCGAN architecture} that expects a square matrix commonly used to represent images. The second is to account for the presence of \textbf{multiple data types} as the proposed \textit{DCGAN} is not designed to handle categorical variables. This sub-section explains how to overcome these issues in detail. Additionally it briefly covers the training objectives used to train CTAB-GAN.

First, each row belonging to the original dataset is transformed as explained in Sec.~\ref{Ch4:data_representation}. Let the size of such a transformed row $r$ be defined as $1 \times T$ where $T$ is the length of each transformed data-row.  Next, the novel \textit{extended conditional vector} (i.e., $ex\_cond$) is sampled as illustrated in section \ref{Ch4:condvec}. Let the size of $ex\_cond$ be $1 \times E$. The extended conditional vector and it's corresponding real data-row are further concatenated to form a vector (i.e., $r \oplus ex\_cond$) of size $1 \times (T+E)$. \\
\\
To deal with \textbf{data compatibility}, each data-row and it's \textit{condition vector} stored as a vector of size $1\times(T+E)$ is wrapped into the closest square matrix of dimensions, i.e. $1\times d \times d$ such that $d$ is the ceiled square root of the data-row dimensionality (i.e., $T+E$). And, unfilled entries of the square matrix are padded with zeros. For example, for a data row with 8 variables, it is converted into a square matrix of dimension $3\times3$ where the last missing entry corresponding to an additional $9^{th}$ column is filled with a zero. 

This square shaped image-like format is then used to define the input layer dimensions of $\mathcal{D}$ to take as input of shape i.e., $1 \times d \times d$ where 1 is the number of channels and $d$ is the height and width, respectively. The generator on the other hand is initialized to take in as input a random noise vector $z$ of arbitrary size $s$ coupled with it's corresponding conditional vector $ex\_cond$ of size $E$ (i.e., $z \oplus ex\_cond$ of size $(s+E)\times1\times1$) to output a square matrix of shape $1 \times g \times g$ where $g$ is calculated as ceiled square root of $T$ (i.e., the generator is not required to generate data concatenated with conditional vectors).\\
\\
To account for \textbf{multiple data-types}, the output of the generator is converted back into the shape of the original tabular encoding $1 \times T$ after discarding the additional columns gained as a result of converting to a square matrix. Subsequently, the final activation is applied. For the scalar values $\alpha$ for mixed $\&$ continuous variables, a \textit{Tanh} final activation is used. And for one-hot-encodings used for representing the modes (i.e., $\beta$) and the categorical variables (i.e., $\gamma$) (refer to Sec.~\ref{Ch4:data_representation}), the \textit{gumbel softmax activation} function with a temperature parameter of $0.20$ is used. This is based on the work of \cite{ctgan} where the different activations accounts for the difference in data-types (a non-issue for generating images). 

The resulting generated tabular data-row $\hat{r}$ of size $1\times T$ is concatenated with it's corresponding conditional vector $ex\_cond$ of size $1\times E$ (i.e., $\hat{r}\oplus ex\_cond$ of size $1\times(T+E)$). And, this is similarly converted back to a square shape of size $1 \times d \times d$ to be passed to the discriminator. \\
\\
Finally, to account for the training objectives, let $\mathcal{L}^{D}_{orig}$ and $\mathcal{L}^{G}_{orig}$ denote the original GAN loss functions from~\cite{gan} described in Sec.~\ref{Ch4:dcgan} to train the discriminator $\mathcal{D}$ and generator $\mathcal{G}$, respectively. Furthermore, for the generator (i.e., $\mathcal{G}$) the complete training objective is the combination of the classification, information and generator losses ( refer to Sec.~\ref{Ch4:TB}). Thus, the training objective can be formally expressed as: $\mathcal{L}^{G}=\mathcal{L}^{G}_{orig}+\mathcal{L}_{class}^{G}+\mathcal{L}_{info}^{G} + \mathcal{L}_{generator}^{G}$, while for $\mathcal{D}$ it remains unchanged, i.e. $\mathcal{L}^{D}_{orig}$. \\
\\
Lastly, it is important to note that for utilising the classifier module, the conditional vectors are not concatenated to either real or generated samples. Furthermore, the generated data is not converted back into a square form after applying the final activation and is used as is. In this way, the classifier takes as input the tabular encoded data representation of real/synthetic data expressed in Sec.~\ref{Ch4:data_representation}.

\section{Experimental Analysis}
\label{Ch4:exp}

To show the efficacy of the proposed CTAB-GAN model, the experimental analysis introduced in Sec. \ref{Ch3:EC} is extended to include CTAB-GAN. Hence the same experimental setup is used to compare the performance of CTAB-GAN with respect to the baselines set by the four state-of-the-art GAN generators introduced therein in terms of the resulting ML utility, statistical similarity to the real data, and privacy distance. Additionally, we provide an ablation analysis to highlight the efficacy of the unique components of CTAB-GAN. 

\subsection{Results analysis}
\label{Ch4:Results}

\begin{figure}[htb]
 	\begin{center}
		\subfloat[\centering Adult]{
			\includegraphics[width=0.3\textwidth]{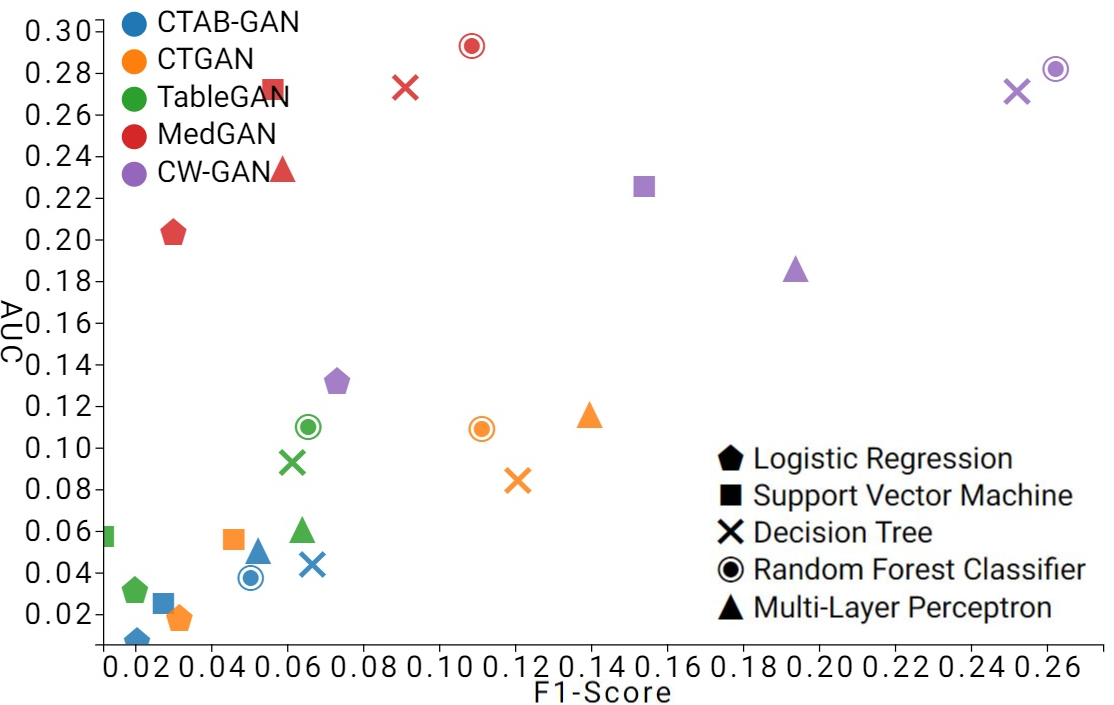}
			\label{fig:result_adult}
		}
	\hspace{\fill}
		\subfloat[\centering Covertype]{
			\includegraphics[width=0.3\textwidth]{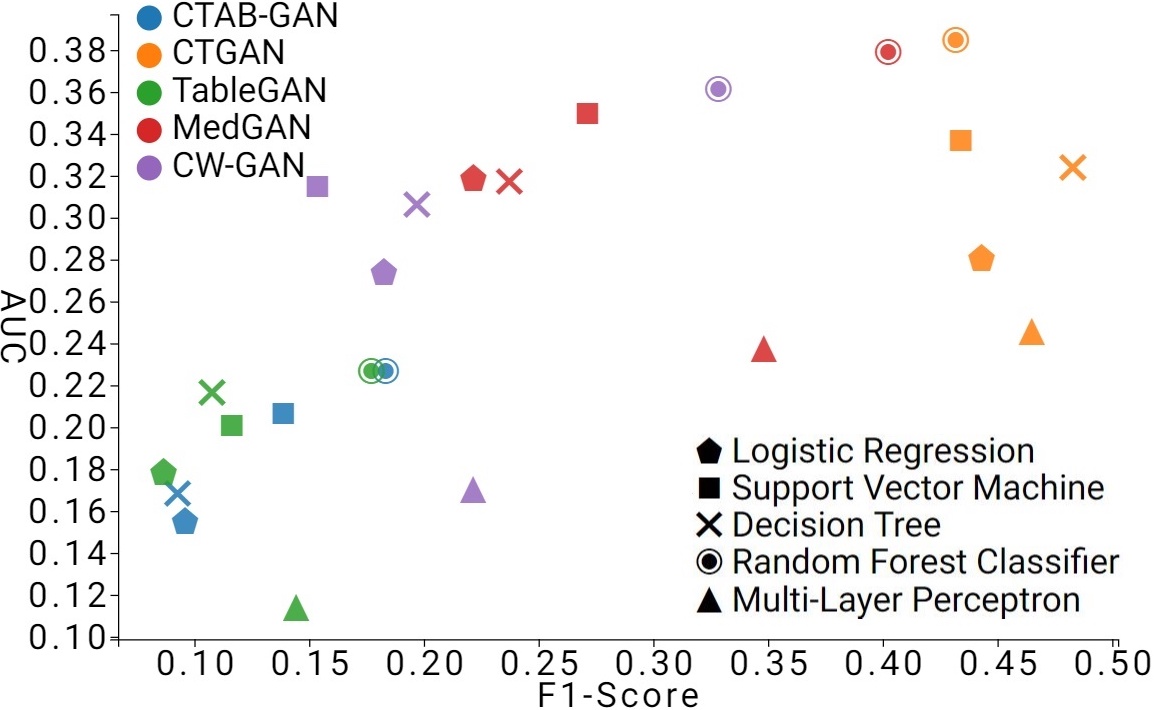}
			\label{fig:result_covtype}
 		}
	\hspace{\fill}
		\subfloat[\centering Credit]{
			\includegraphics[width=0.3\textwidth]{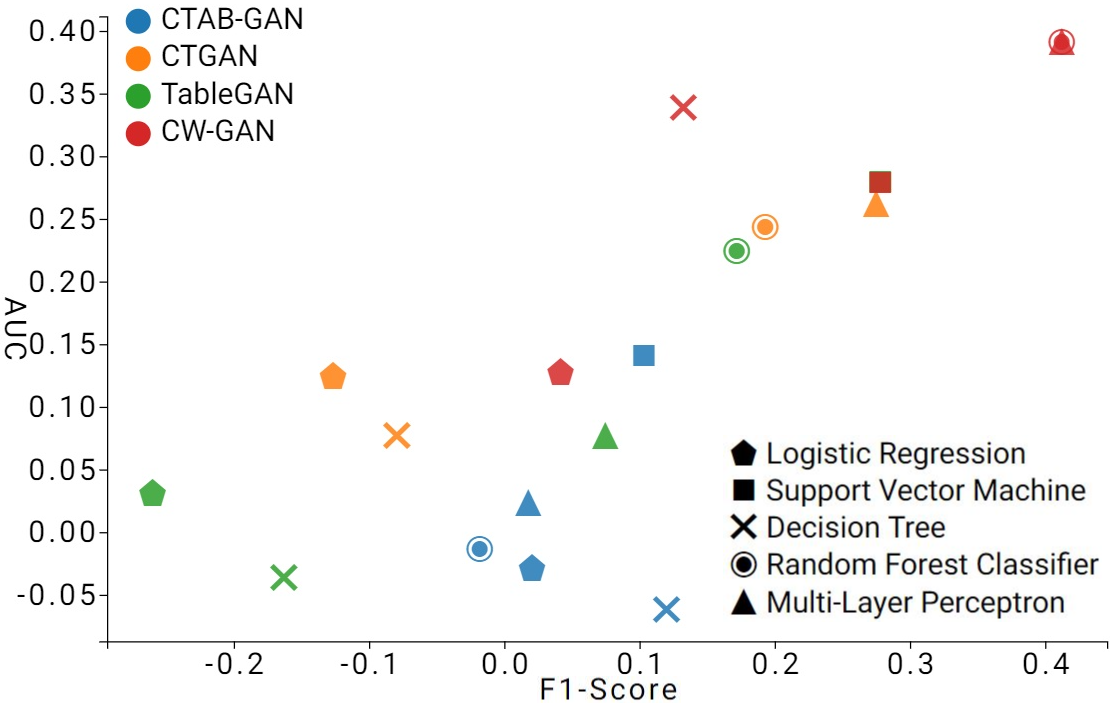}
			\label{fig:result_credit}
		}
	\hspace{\fill}
		\subfloat[\centering Intrusion]{
			\includegraphics[width=0.3\textwidth]{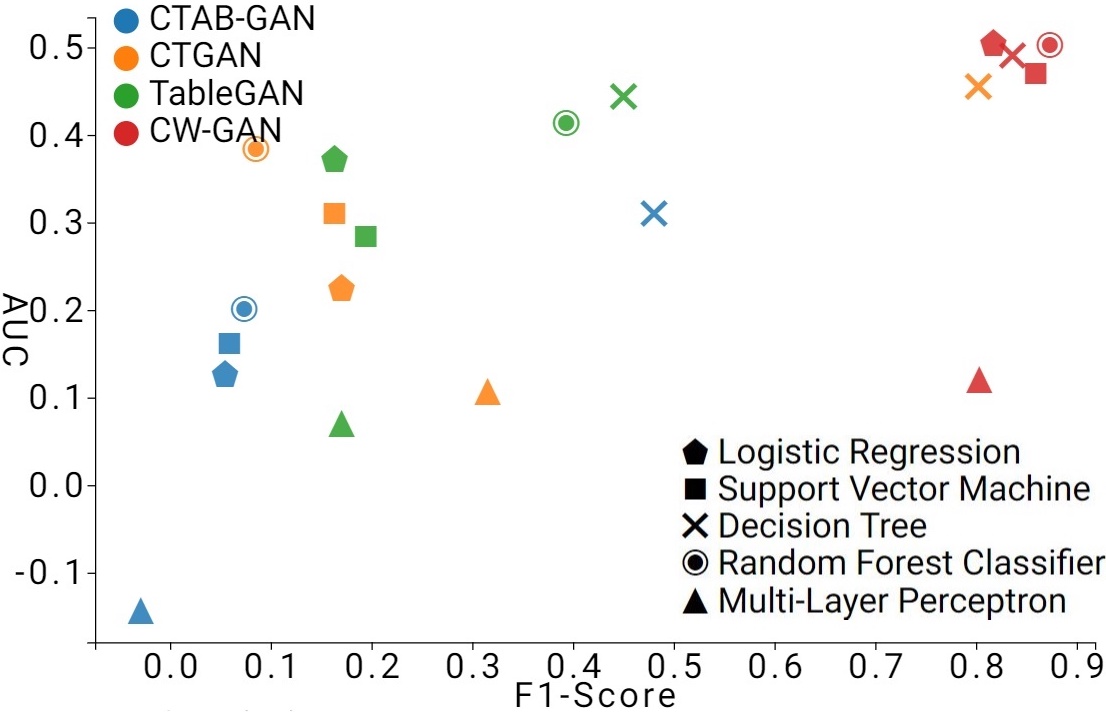}
			\label{fig:result_intrusion}
		}
	\quad
		\subfloat[\centering Loan]{
			\includegraphics[width=0.3\textwidth]{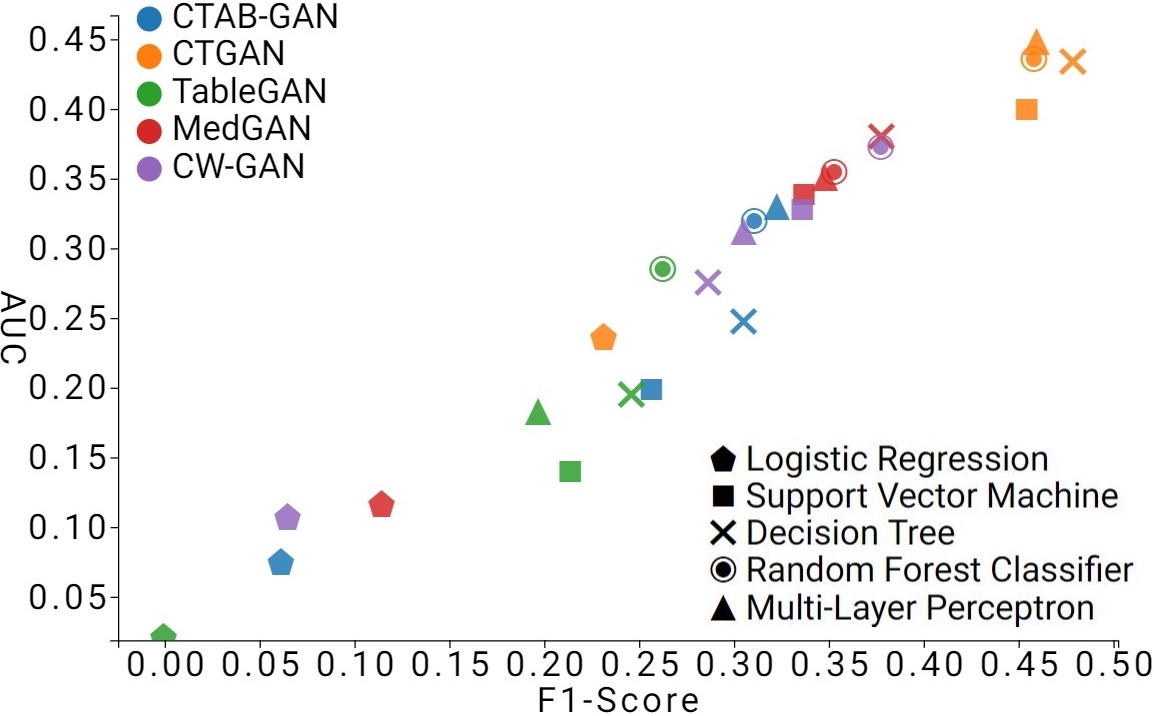}
			\label{fig:result_loan}
		}
		\caption{ML utilities difference (i.e., AUC and F1-score) for five algorithms using five synthetic data generators on all 5 datasets}
		\label{fig:ml_whole}
	\end{center}
\end{figure}

\begin{enumerate}
    \item \textbf{ML Utility}- Tab.~\ref{table:ML_all} shows the averaged ML utility differences between real and synthetic data in terms of accuracy, F1 score, and AUC. A better synthetic dataset is expected to have low differences. It can be seen that CTAB-GAN outperforms all other state-of-the-art methods in terms of Accuracy, F1-score and AUC. Accuracy is the most commonly used classification metric, but to account for imbalanced target variables, the F1-score and AUC are more reliable metrics to evaluate performance. CTAB-GAN largely shortens the AUC difference from 0.169 (best in state-of-the-art) to 0.094.  

\begin{table}[htb]
\centering
\caption{\centering Difference of ML accuracy (\%), F1-score, and AUC between original and synthetic data: average over 5 different datasets and 3 replications.}
\begin{tabular}{|c|c|c|c|}
\hline
\textbf{Method} & \textbf{Accuracy} & \textbf{F1-score}  & \textbf{AUC} \\
\hline
\small{CTAB-GAN}    & \textbf{8.90\%} & \textbf{0.107} &   \textbf{0.094}  \\
\small{CTGAN}    &21.51\% & 0.274 &   0.253 \\
\small{TableGAN} &11.40\%  & 0.130 &   0.169  \\
\small{MedGAN}    & 14.11\%&  0.282&   0.285 \\
\small{CW-GAN}    &  20.06\%& 0.354 &  0.299 \\
\hline
\end{tabular}
\label{table:ML_all}
\end{table}

To obtain a better understanding, Fig.~\ref{fig:ml_whole} plots the (F1-score-x axis, AUC-y-axis) for all 5 ML models for all datasets. 

Fig.~\ref{fig:ml_whole}(a,b $\&$ c) show that for Adult, Covtype and Credit datasets, the results of CTAB-GAN and TableGAN are largely similar and clearly better than the rest. This is due to a more stable \textit{DCGAN architecture} that trains reliably and therefore generates high utility datasets. 

Fig.~\ref{fig:ml_whole}(d) shows that for the Intrusion dataset, CTAB-GAN largely outperforms all others across all ML models used for evaluation. This can be explained by the use of the \textit{conditional GAN architecture} that helps deal with imbalanced variables and the added \textit{information loss} which greatly helps stabilize training (refer to Sec.~\ref{Ch4:ablation}). 

Fig.~\ref{fig:ml_whole}(e) however shows that TableGAN outperforms CTAB-GAN on the loan dataset. The Loan dataset is significantly smaller than the other four. Therefore, we find that the encoding method in CTAB-GAN which works well for complex cases also increases the dimensionality of the input data. This results in a failure to converge to a better optimum for smaller datasets. Whereas TableGAN's encoding doesn't lead to an increase in the dimensionality of the raw data as \textit{categorical variables} are simply treated as \textit{continuous} and no \textit{MSN} (refer to Sec.~\ref{Ch4:msn}) is used. Thus, this leads to a simpler representation making it easier for the TableGAN model to learn effectively from smaller datasets.

\item \textbf{Statistical similarity}- Statistical similarity results are reported in Tab.~\ref{table:SS_all}. CTAB-GAN stands out again across all comparisons. 
 
For \textit{categorical variables} (i.e. average JSD), CTAB-GAN outperforms CTGAN and TableGAN by 13.5\% and 28.4\%. Although both CTGAN and CTAB-GAN rely on conditional GAN and training-by-sampling to deal with categorical imbalance, the addition of a superior \textit{DCGAN neural network architecture} and the additional loss terms for the generator such as the \textit{classification} and \textit{information losses} enable CTAB-GAN to outperform it's predecessor.

For \textit{continuous variables} (i.e. average WD), CTAB-GAN benefits from the design of the \textit{mixed-encoder} to deal with \textit{mixed data variables}. Moreover, the use of an \textit{extended conditional vector} helps to better produce \textit{skewed multi-modal numerical distributions}. And, the use of the \textit{log-transform} allows to better capture \textit{long-tailed distributions} (refer to Sec.~\ref{Ch4:motivation_response}). It is worth pointing out that the average WD column shows some extreme numbers such as 46257 and 238155 comparing to 1197 of CTAB-GAN due to these algorithms generating extremely large values for long tail variables. 

Besides divergence and distance, CTAB-GAN's synthetic data also maintains the best \textit{correlation}. The \textit{extended conditional vector} allows the generator to produce samples conditioned even on a given \textit{VGM mode} for \textit{continuous variables}. This increases the capacity to learn the \textit{conditional distribution} for \textit{continuous variables} and hence leads to an improvement in the overall feature interactions captured by the model.  

\begin{table}[htb]
\centering
\caption{\centering Statistical similarity: three measures averaged over 5 datasets and three repetitions.}
\begin{tabular}{|c|c|c|c|}
\hline
\textbf{Method} & \textbf{Avg JSD} & \textbf{Avg WD}  & \textbf{Diff. Corr.} \\
\hline
\small{CTAB-GAN}   & \textbf{0.062}&  \textbf{1197}  &  \textbf{2.09} \\
\small{CTGAN}    & 0.0704&  1769  & 2.73  \\
\small{TableGAN} & 0.0796& 2117   &  2.30 \\
\small{MedGAN}   & 0.2135& 46257  &   5.48   \\
\small{CW-GAN}   & 0.1318& 238155 &  5.82 \\
\hline
\end{tabular}
\label{table:SS_all}
\end{table}

\item \textbf{Privacy preservability}- The privacy results are shown in Tab.~\ref{table:PP_all}. It can be seen that the DCR and NNDR between real and synthetic data all indicate that generation from TableGAN has the shortest distance to real data (highest privacy risk). 

Moreover, as we use distance-based algorithms to give an overview on privacy, the evaluation of privacy is relative to the utility. This is because, on the one hand, if the distance between real and synthetic data is too large, it simply means that the quality of generated data is poor. On the other hand, if the distance between real and synthetic data is too small, it simply means that there is a risk to reveal sensitive information from the training data.  

Thus, the algorithm which allows for greater distances between real and synthetic data under equivalent ML utility and statistical similarity data should be considered. In that case, CTAB-GAN not only outperforms TableGAN in ML utility and statistic similarity, but also in all privacy preservability metrics by 11.6\% and 4.5\% for DCR and NNDR, respectively. 

\begin{table}[htb]
\centering
\caption{\centering Privacy impact: between real and synthetic data (R\&S) and within real data (R) and synthetic data (S).}
\begin{tabular}{|c|c|c|c|c|c|c|}
\hline
\multirow{2}{*}{\textbf{Model}} & \multicolumn{3}{c|}{\textbf{DCR}} & \multicolumn{3}{c|}{\textbf{NNDR}} \\
\cline{2-7}
 & \textbf{R\&S} & \textbf{R}  & \textbf{S} & \textbf{R\&S} & \textbf{R}  & \textbf{S}\\

\hline
CTAB-GAN    & 1.118 &  0.428& 0.937 & 0.713 & 0.414 &0.591\\
CTGAN    & 1.517 &  0.428& 1.026 & 0.763 & 0.414 &0.624 \\
TableGAN & 0.988 &  0.428& 0.920 & 0.681 & 0.414 &0.632\\
MedGAN   & 1.918 &  0.428& 0.254 & 0.871 & 0.414 &0.393 \\
CW-GAN   & 2.197 &  0.428& 1.124 & 0.847 & 0.414 &0.675\\
\hline
\end{tabular}
\label{table:PP_all}
\end{table}

\end{enumerate}

\subsection{Ablation analysis}
\label{Ch4:ablation}

To illustrate the efficiency of each strategy we implement an ablation study which cuts off the different components of CTAB-GAN one by one: 

\begin{enumerate}
    \item \textbf{w/o $\mathcal{C}$}- In this experiment, Classifier $\mathcal{C}$ and the corresponding classification loss for Generator $\mathcal{G}$ are taken away from CTAB-GAN
    \item \textbf{w/o I. loss} (information loss)- In this experiment, we remove information loss from CTAB-GAN
    \item \textbf{w/o MSN}- In this case, we substitute the mode specific normalization based on VGM for continuous variables with min-max normalization and use simple one-hot encoding for categorical variables. Here the conditional vector is the same as for CTGAN
    \item \textbf{w/o LT}- (long tail). In this experiment, long tail treatment is no longer applied. This only affects datasets with long tailed columns, i.e. Credit and Intrusion.
\end{enumerate}

The results are compared with the baseline implementing all strategies. All experiments are repeated 3 times, and results are evaluated on the same 5 machine learning algorithms introduced in Sec.~\ref{Ch3:ml_efficacy}. We report the F1-score difference between CTAB-GAN and each above-mentioned experiments where the test datasets and evaluation flow are the same as shown in Sec.~\ref{Ch3:EC} and Sec.~\ref{Ch3:metrics}. Tab.~\ref{table:ablation} shows the results.

\begin{table}[htb]
\centering
\caption{\centering F1-score difference to CTAB-GAN. CTAB-GAN column reports the absolute averaged F1-score as baseline.}
\label{table:ablation}
\begin{tabular}{ |c|c|c|c|c|c|}
\hline
\textbf{Dataset} & \textbf{CTAB-GAN} & \textbf{w/o $\mathcal{C}$} & \textbf{w/o I. Loss}  & \textbf{w/o MSN} & \textbf{w/o LT} \\
\hline
Adult  & 0.704 &  -0.01&	-0.037 & -0.05  & -   \\
Covertype &  0.532 & -0.018 & -0.184& -0.118 & - \\
Credit & 0.710 & +0.011& -0.177& +0.06 &  0.00 \\
Intrusion & 0.842&-0.031&-0.437&+0.003 & -0.074 \\
Loan & 0.803 &-0.044&+0.028&+0.013 & -  \\
\hline
\end{tabular}
\end{table}

Each part of CTAB-GAN has different impacts on different datasets as follows:

\begin{enumerate}
    \item \textbf{w/o $\mathcal{C}$}- has a negative impact for all datasets except Credit. Since Credit has only 30 continuous variables and one target variable, the semantic check can not be very effective.
    \item \textbf{w/o I. loss}- has a positive impact for Loan, but results degenerate for all other datasets. It can even make the model especially unusable for Intrusion. This shows that the information loss is worse for smaller datasets and beneficial for larger datasets.
    \item \textbf{w/o MSN}- performs worse for Covertype and Adult, has little impact for Intrusion and provides better results for the Credit and Loan datasets than the original CTAB-GAN. This is because out of 30 continuous variables in the Credit dataset, 28 are nearly single mode Gaussian distributed. Thus, the initialized high number of modes, i.e. 10, for each continuous variable (same setting as in CTGAN) degrades the estimation quality.  Likewise, for the Loan dataset, the MSN encoding increases the input data dimensionality greatly thereby increasing the difficulty of learning from smaller sized dataset such as Loan.
    \item \textbf{w/o LT}- has the biggest impact on Intrusion, since it contains 2 long tail columns which are seemingly important predictors for the target column. For Credit, the  influence is limited. Even if the long tail treatment fits the \textit{amount} column well (see Sec.~\ref{Ch4:motivation_response}), this variable doesn't seem to be a strong predictor for the target column. 
\end{enumerate}

In general, averaging the column values across all ablation tests results in a negative impact for the performance which justifies our design choices for CTAB-GAN.

 \subsection{Results for Motivation Cases}
 \label{Ch4:motivation_response}
 
After reviewing all the metrics, let us recall the three motivation cases from Sec.~\ref{Ch3:Challenges}. 

\begin{figure}[htb]
	\begin{center}
		\subfloat[\centering Mortgage in Loan dataset~\cite{kaggleloan}]{
			\includegraphics[width=0.33\textwidth]{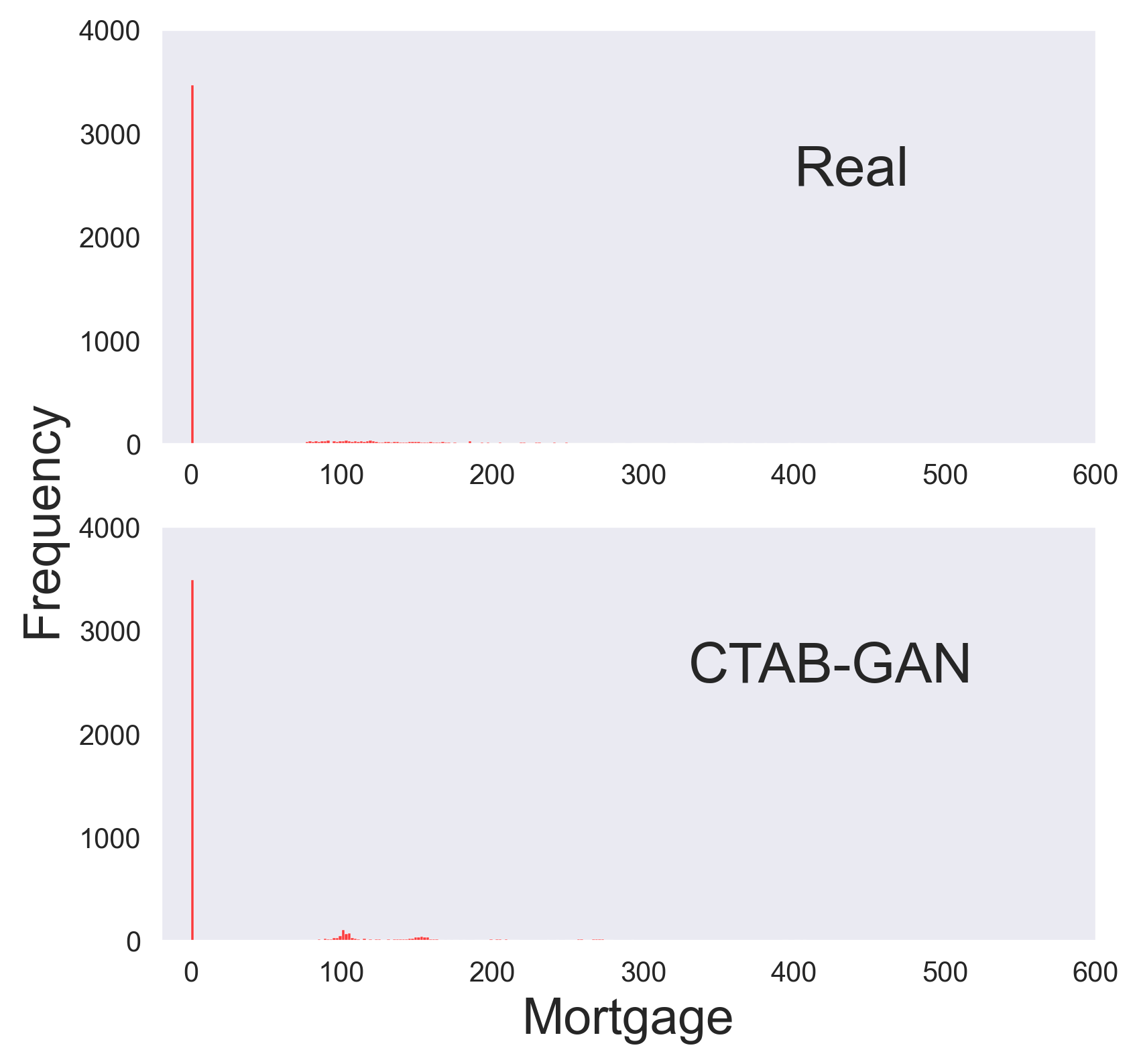}
			\label{fig:ctabgan_bimodal}
		}
		\subfloat[\centering Amount in Credit dataset~\cite{kagglecredit}]{
			\includegraphics[width=0.3\textwidth]{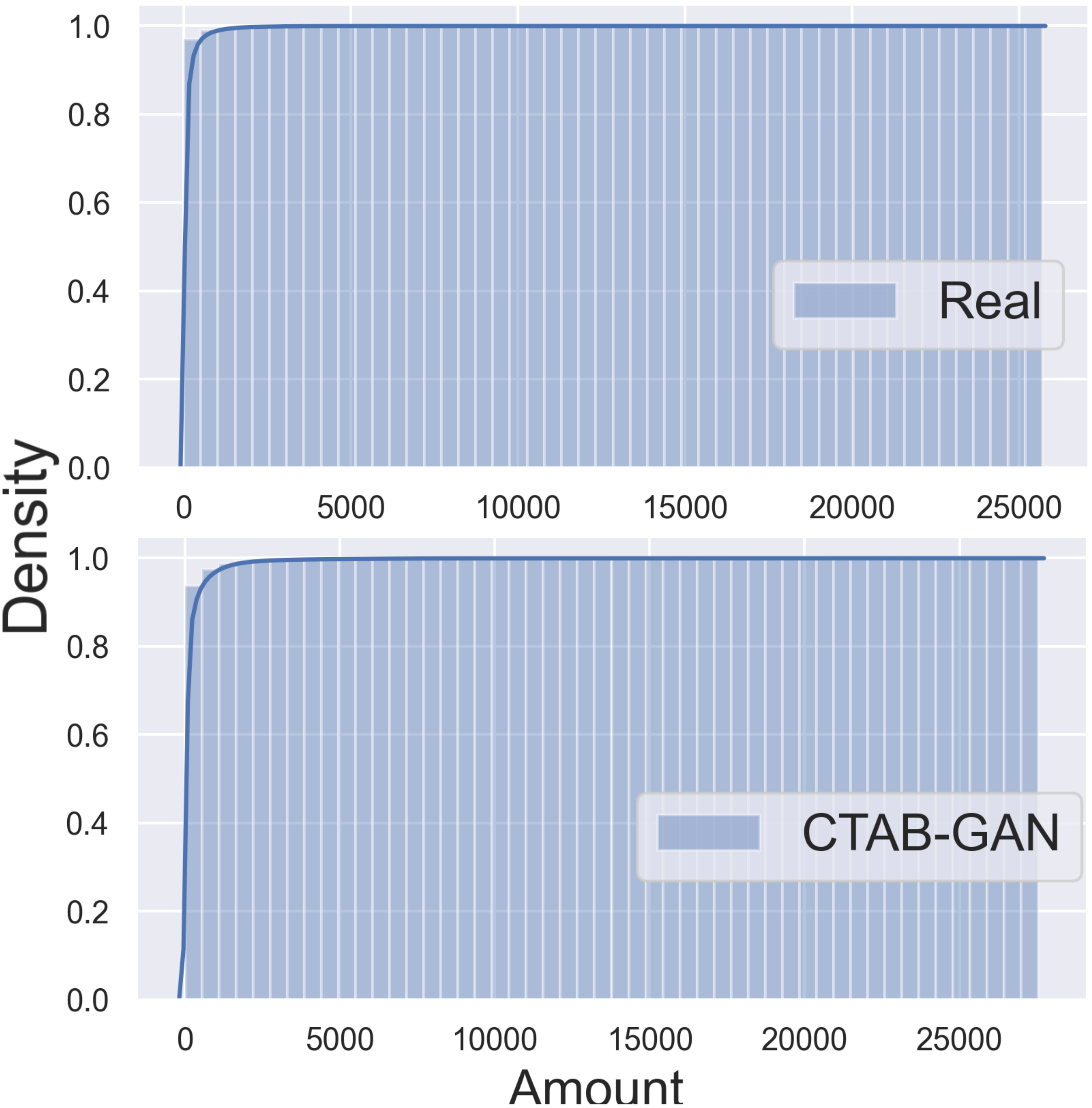}
			\label{fig:longtail_result}
		}
		\subfloat[\centering Hours-per-week in Adult dataset~\cite{UCIdataset}]{
			\includegraphics[width=0.335\textwidth]{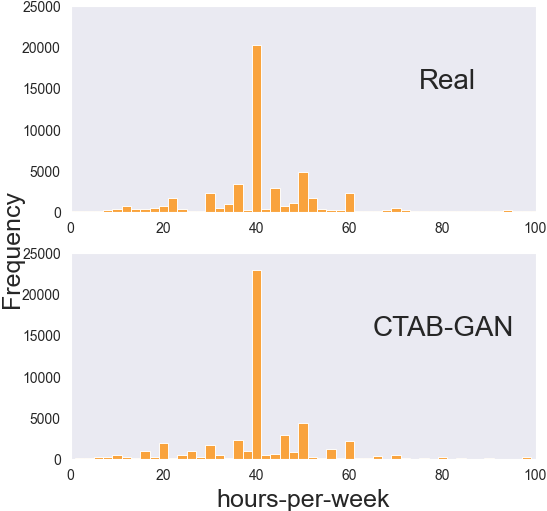}
			\label{fig:gmm_result}
		}
		\caption{ \centering Challenges of modeling industrial dataset using existing GAN-based table generator: (a) Mixed data type, (b) long tail distribution, and (c) Skewed multi-modal data} 
		\label{fig:motivationcases_response}
 	\end{center}
\end{figure}

\begin{enumerate}
    \item \textbf{Mixed data type variables}- Fig.~\ref{fig:motivationcases_response}(a) compares the real and CTAB-GAN generated data for variable \textit{Mortgage} in the Loan dataset. CTAB-GAN encodes this variable as mixed data-type. We can see that CTAB-GAN generates clear 0 values and the frequency is similar as in real distribution. Therefore, this is a result of using the mixed encoder combined with the extended conditional vector to control the sampling of the categorical component to correspond to the original data with greater similarity. 
    \item \textbf{Long tail distributions}- Fig.~\ref{fig:motivationcases_response}(b) compares the cumulative frequency graph for the \textit{Amount} variable in Credit. This variable is a typical long tail distribution. One can see that CTAB-GAN perfectly recovers the real distribution. Due to log-transform data pre-processsing, CTAB-GAN learns this structure significantly better than the state-of-the-art methods shown in Fig.~\ref{fig:motivationcases}(b).
    \item \textbf{Skewed multi-mode continuous variables}-Fig.~\ref{fig:motivationcases_response}(c) compares the frequency distribution for the continuous variable \textit{Hours-per-week} from Adult. Except the dominant peak at 40, there are many side peaks. Fig.~\ref{fig:motivationcases}(c), shows that TableGAN, CW-GAN and MedGAN struggle since they can learn only a simple Gaussian distribution due to the lack of any special treatment for continuous variables. CTGAN, which also use VGM, can detect other modes. However, CTGAN is not as good as CTAB-GAN. The reason is that CTGAN lacks the mode of continuous variables in the conditional vector. By incorporating the  mode of continuous variables into conditional vector, we can apply the training-by-sample and logarithm frequency also to modes. This gives the mode with less weight more chance to appear in the training and avoids the mode collapse. 
\end{enumerate}

\section{Conclusion}
\label{Ch4:Conclusion}

Motivated by the importance of data sharing and fulfillment of governmental regulations, we propose CTAB-GAN -- a novel conditional GAN based tabular data generator. CTAB-GAN advances beyond the prior state-of-the-art methods by modeling mixed data-type variables and provides strong generation capabilities for long-tailed continuous variables and continuous variables with complex distributions.

To such ends, the core features of CTAB-GAN include (i) introduction of the classification and information loss into the conditional DCGAN, (ii) effective data encoding for mixed data-type variables, and (iii) a novel construction of conditional vectors. 

We exhaustively evaluate CTAB-GAN against four tabular data generators on a wide range of metrics, namely ML utilities, statistical similarity and privacy preservation. The results show that the synthetic data of CTAB-GAN results into  high utilities, high similarity and reasonable privacy guarantee, compared to existing state-of-the-art techniques. The improvement on complex datasets is up to 17\% in accuracy comparing to all state-of-the-art algorithms. 


%% file: Content/Chapters/5_DP_CTABGAN.tex
\chapter{Differential Privacy for Tabular Data Generators }\label{ch5}

\section{Introduction}
The previous chapters illustrated the efficacy of tabular GANs for learning the training data distributions and generating high utility synthetic datasets. However, utilising privacy sensitive real datasets to train tabular GANs poses a range of privacy issues. Recent studies have shown that GANs may fall prey to membership and attribute inference attacks which greatly endanger the personal information present in the real training data~\cite{gan_leak,priv_mirage}. Therefore, it is imperative to safeguard the training of tabular GANs such that it remains protected against malicious privacy attacks to ensure that synthetic data can be stored and shared across different parties without harm.  

The limited existing work~\cite{pategan,long2019scalable,torkzadehmahani2019dp,torfi2020differentially} rely on Differential Privacy (DP)~\cite{dwork2008differential} for training tabular GANs in a privacy preserving manner. DP is a mathematical framework that provides theoretical guarantees bounding the statistical difference between any resulting tabular GAN model trained regardless of the existence of any particular individual's information in the original training dataset. Typically, this is achieved by (i) clipping the gradients for bounding the sensitivity and (ii) injecting noise while updating the parameters of a network during back-propagation~\cite{backprop}. However, the main challenge found in prior work is to calibrate the training of differential private tabular GANs so as to maintain the utility of synthetic datasets for analysis while providing strict theoretical privacy guarantees. Moreover, the existing literature rarely investigates the empirical robustness of their differential private GANs against privacy attacks.

In this chapter, two variants of differential private CTAB-GAN are proposed based on the ideas presented in prior work, most notably, the work done by the authors of DP-WGAN~\cite{xie2018differentially} and GS-WGAN~\cite{chen2020gs}. Furthermore, a rigorous empirical evaluation is conducted to investigate the usefulness of differential private tabular GANs in terms of their utility for analysis given their constraints to preserve privacy especially against malicious privacy attacks such as the membership and attribute inference attacks. 

The rest of this chapter is organized as follows:  the two main approaches used to employ differential privacy in CTAB-GAN are elucidated in Sec.~\ref{Ch5:dpctabgan}. Then in Sec.~\ref{Ch5:EA}, a rigorous empirical examination of DP-CTABGAN is provided. Finally, Sec.~\ref{Ch5:Conclusion} ends the chapter with a succinct summary of the results and provides directions for further research.

\section{DP-CTABGAN}
\label{Ch5:dpctabgan}

DP-CTABGAN is a novel approach to generate tabular datasets with strong DP guarantees. It utilizes the DP-SGD~\cite{abadi2016deep} framework introduced by \cite{abadi2016deep} and the subsampled RDP moments accountant technique~\cite{mironov2017renyi,wang2019subsampled} to preserve privacy and account for the cost, respectively. In addition, it makes use of the wasserstein loss with gradient penalty~\cite{gulrajani2017improved} to effectively bound the gradient norms with an analytically derived optimal clipping value as shown in the work of \cite{chen2020gan}. Therefore, the rest of this section is organised as follows: First, Sec.~\ref{Ch5:WGAN_real} highlights the updated training objective of DP-CTABGAN. Next, Sec.~\ref{Ch5:DPD} $\&$ Sec.~\ref{Ch5:DPG} presents two variants of DP-CTABGAN. Sec.~\ref{Ch5:DPD} details the implementation and privacy analysis for training the discriminator network with DP guarantees whereas in Sec.~\ref{Ch5:DPG}, the generator network is described. Both approaches are studied to obtain the most optimal configuration for training DP-CTABGAN.

\subsection{Wasserstein Loss with Gradient Penalty~\cite{gulrajani2017improved}}
\label{Ch5:WGAN_real}

One of the biggest challenges with using DP-SGD is tuning the clipping parameter, $C$, for bounding the gradient norms. Since clipping greatly degrades the information stored in the original gradients~\cite{chen2020gs}, choosing an optimal clipping value that does not significantly impact utility is crucial. 

However, tuning the clipping parameter is laborious as the optimal value fluctuates depending on network hyperparameters (i.e model architecture, learning rate)~\cite{abadi2016deep}. Therefore, inspired by the work of \cite{chen2020gs}, the wasserstein loss with gradient penalty~\cite{gulrajani2017improved} (refer to Sec.~\ref{Ch5:WGAN}) is chosen as a suitable loss function for training both variants of DP-CTABGAN.

The gradient penalty term is especially useful as it ensures that the discriminator generates bounded gradient norms which are close to 1 under real and generated distributions. Therefore, an optimal clipping threshold of $C=1$ is obtained analytically avoiding an intensive hyper-parameter search thereby better preserving the information stored in gradients after clipping.

Note that the prior implementation of CTAB-GAN made use of batch normalization to help improve the flow of gradients in both the generator and the discriminator network. However, with the updated gradient penalty training objective which penalizes the gradients for each input data point independently, it is no longer valid. Therefore, \cite{gulrajani2017improved} recommends utilising layer normalization~\cite{ba2016layer} as a drop-in replacement for batch normalisation as it doesn't induce any correlations between data points. And, it was found that layer normalisation significantly improved the flow of gradient information during training based on preliminary experiments.

Additionally, utilising a simple linear interpolation between real and synthetic data points for computing the gradient penalty relies on the assumption that data points form a uniformly distributed hypercube. Since this assumption may not always hold in practice, spherical interpolates~\cite{shoemake1985animating} are used in this work for accounting the possible curvature of the latent space. And, it was found to yield better data utility in preliminary experiments. 

\subsection{DP-Discriminator}
\label{Ch5:DPD}
In the first variant, DP-CTABGAN trains the discriminator using differential private-SGD as outlined in algorithm~\ref{Ch5:DPSGD-Algo} where the total number of iterations $T$ is determined based on the total privacy budget $(\epsilon$,$\delta)$. Thus, to compute the number of iterations, the privacy budget spent for every iteration must be bounded and accumulated over training iterations $T$. The subsampled RDP analytical moments accountant technique~\cite{wang2019subsampled} is used for this purpose. The theoretical analysis of the privacy cost is presented below: 
\\
\\
\textbf{Theorem 5.2.1} Each discriminator update satisfies $(\lambda,2B\lambda/\sigma^{2})$-RDP where B is the batch size. 
\\
\emph{Proof.} Let $f=clip({\bar{g}_D},C)$ be the clipped gradient of the discriminator before adding noise. The sensitivity is derived via the triangle inequality:
\begin{equation}
    \Delta_{2}f = \max_{S,S'}||f(S)-f(S')||_{2} \leq 2C
\end{equation}

Since $C=1$ as a consequence of the wasserstein loss with gradient penalty~\cite{gulrajani2017improved} and by using definition 2.2.3 in Sec.~\ref{Ch5:background}, the DP-SGD procedure denoted as $\mathcal{M}_{\sigma,C}$ parameterized by noise scale $\sigma$ and clipping parameter $C$ may be represented as being $(\lambda,2\lambda/\sigma^{2})$-RDP. 

Furthermore, each discriminator update for a batch of real data points $\{x_i,..,x_B\}$ can be represented as 

\begin{equation}
    \Tilde{g}_D = \frac{1}{B}\sum_{i=1}^{B}\mathcal{M}_{\sigma,C}(\nabla_{\theta_D}\mathcal{L}_{D}(\theta_D,x_i))
\end{equation}

where $\tilde{g}_{D}$ and $\theta_{D}$ represents the perturbed gradients and the weights of the discriminator network, respectively. This may be regarded as a composition of B Gaussian mechanisms. And so, by using theorem 2.2.1 in Sec.~\ref{Ch5:background}, the privacy cost for a single gradient update step for the discriminator can be expressed as $(\lambda,\sum_{i=1}^{B}2\lambda/\sigma^{2})$ or equivalently $(\lambda,2B\lambda/\sigma^{2})$. \tiny\qedsymbol

\normalsize Note that $\mathcal{M}_{\sigma,C}$ is only applied for those gradients that are computed with respect to the real training dataset~\cite{abadi2016deep,zhang2018differentially}. Hence, the gradients computed with respect to the synthetic data and the gradient penalty term are left undisturbed. 

Next, to further amplify the privacy protection of the discriminator, theorem 2.2.3 defined in Sec.~\ref{Ch5:background} is used where the subsampling rate is defined as $\gamma =B/N$ where $B$ is the batch size and $N$ is the size of the training dataset. Intuitively, subsampling adds another layer of randomness and enhances privacy by decreasing the chances of leaking information about particular individuals who are not included in any given subsample of the dataset. 

Lastly, it is worth mentioning that the wasserstein loss with gradient penalty~\cite{gulrajani2017improved} training objective has one major pitfall with respect to the privacy cost. This is because, it encourages the use of a stronger discriminator network to provide more meaningful gradient updates to the generator. This requires performing multiple updates to the discriminator for each corresponding update to the generator leading to a faster consumption of the overall privacy budget.

\subsection{DP-Generator}
\label{Ch5:DPG}

In the second variant, DP-CTABGAN trains the generator network with DP guarantees. To do so, the gradients flowing from the discriminator and classifier networks (i.e., $g_{G}^{Disc}$ $\&$ $g_{G}^{Class}$) which interact with the original training data are selectively perturbed (i.e., $\tilde{g}_{G}^{Disc}$ $\&$ $\tilde{g}_{G}^{Class}$) via the familiar DP-SGD procedure represented as a randomized mechanism $\mathcal{M}_{\sigma,C}$ parameterized by noise scale $\sigma$ and clipping parameter $C$ for updating the generator's weights (i.e., $\theta_{G}$), as shown in the Fig~\ref{fig:Ch5Gen} below. The selective perturbation of the gradients is necessary as the combined training objective of the generator i.e., classification, information loss and generator losses (refer to Sec.~\ref{Ch4:TP}) doesn't entirely depend on the original training data. As an example, consider the generator loss which is only used to ensure that the generated data exactly matches the constraint given by the conditional vector sampled randomly during training and as a result, is independent of the real training data, itself.

\begin{figure}[htb]
    \centering
    \includegraphics[scale=0.1]{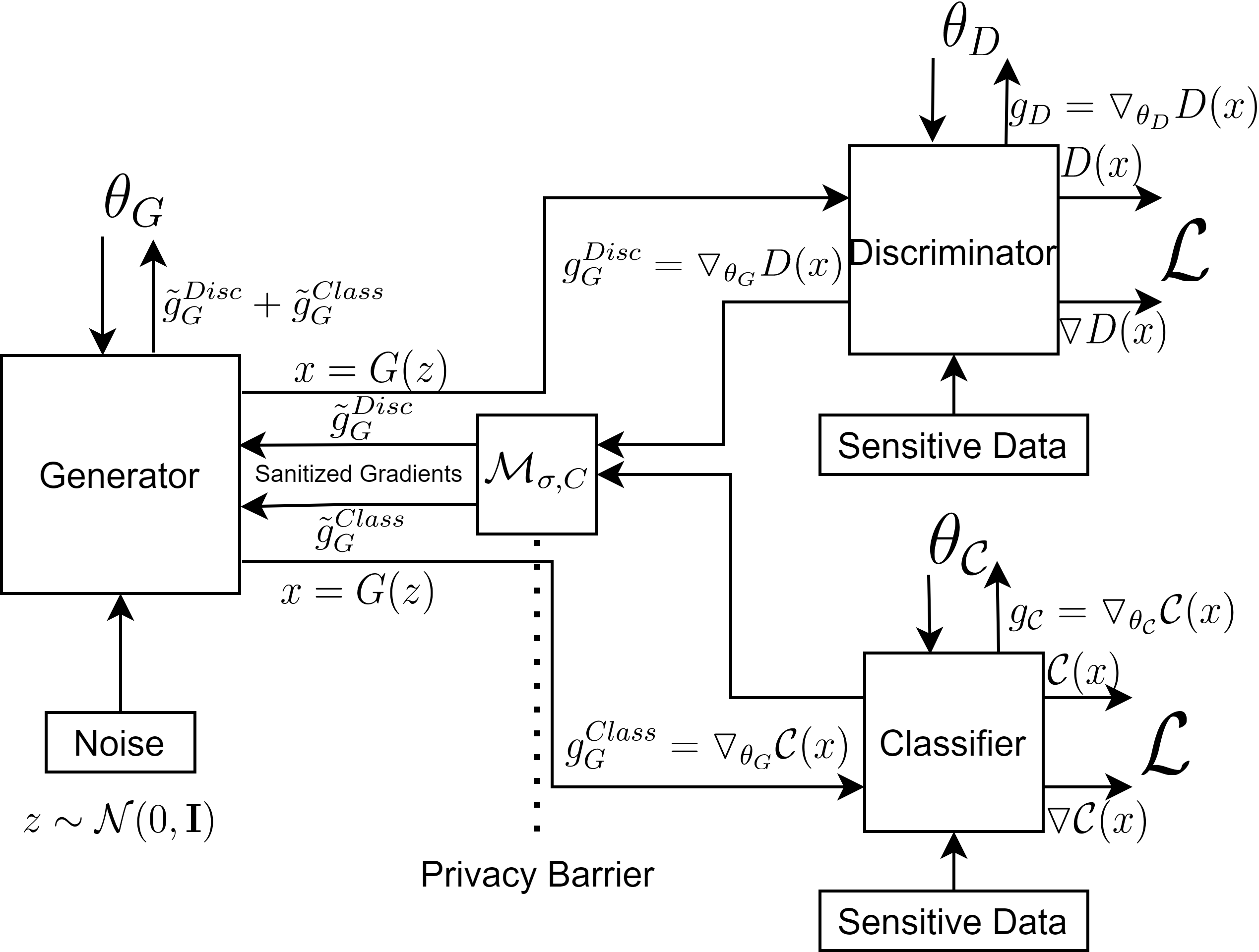}
    \caption{\centering Privacy Preserving Generator Training where $G$,$D$ and $\mathcal{C}$ denote the generator, discriminator and classifier networks with weights $\theta_G$, $\theta_D$ and $\theta_{\mathcal{C}}$, respectively. 
}
    \label{fig:Ch5Gen}
\end{figure}

With this in mind, the privacy analysis for training the generator via DP-SGD~\cite{abadi2016deep} utilizing the aforementioned subsampled RDP moments accountant~\cite{wang2019subsampled} is presented.
\\
\\
\textbf{Theorem 5.2.2} Each generator update satisfies $(\lambda,6B\lambda/\sigma^{2})$-RDP where B is the batch size. 
\\
\emph{Proof} Let $f_{Disc}=clip({\bar{g}_G^{Disc}},C)$ be the clipped gradient of the generator computed with respect to $\mathcal{L}_{G}$ before adding noise. The sensitivity is derived via the triangle inequality:
\begin{equation}
    \Delta_{2}f_{Disc} = \max_{S,S'}||f_{Disc}(S)-f_{Disc}(S')||_{2} \leq 2C
\end{equation}

Since $C=1$ as before and by using definition 2.2.3 in Sec.~\ref{Ch5:background}, the randomized mechanism $\mathcal{M}_{\sigma,C}$ may similarly be represented as being $(\lambda,2\lambda/\sigma^{2})$-RDP. 

However, due to the addition of the information loss denoted as $\mathcal{L}_{I}$, the generator requires an additional fetch of gradients from the discriminator (i.e., $g_G^{Disc}$) computed with respect to $\mathcal{L}_{I}$ which in turn doubles the number of times $\mathcal{M}_{\sigma,C}$ is applied. Note that the sensitivity remains the same leading to an identical privacy cost (i.e., $(\lambda,2\lambda/\sigma^{2})$-RDP).

Likewise for the classifier loss expressed as $\mathcal{L}_{C}$, let $f_{Class}=clip({\bar{g}_G^{Class}},C)$ be the clipped gradient of the generator back-propagated from the classifier before adding noise. The sensitivity is similarly derived via the triangle inequality:

\begin{equation}
    \Delta_{2}f_{Class} = \max_{S,S'}||f_{Class}(S)-f_{Class}(S')||_{2} \leq 2C
\end{equation}

For ease of derivation, the clipping parameter for the classifier module is also, $C=1$. Thus, by using definition 2.2.3 in Sec.~\ref{Ch5:background} once again, $\mathcal{M}_{\sigma,C}$ is $(\lambda,2\lambda/\sigma^{2})$-RDP. 


Thus, to do a single update of the generator's weights $\theta_{G}$, the randomized mechanism $\mathcal{M}_{\sigma,C}$ is first applied twice for the discriminator network and once more for the classifier network with a fixed privacy cost of $(\lambda,2\lambda/\sigma^{2})$-RDP. Formally, this can be expressed as

\begin{equation}
    \tilde{g}_G = \sum_{i=1}^{L} \mathcal{M}_{\sigma,C}(\nabla_{\theta_G}\mathcal{L}_{i}(\theta_G))
\end{equation}
 
where $L$ represents the set of losses for which the gradients are computed (i.e., \{${\mathcal{L}_{G}, \mathcal{L}_{I}, \mathcal{L}_{C}}$\}) and $\tilde{g}_{G}$ $\&$ $\theta_{G}$ represents the perturbed gradients and the weights of the generator network, respectively. This sequence can once again be interpreted as a composition of Gaussian mechanisms which allows the use of theorem 2.2.1 defined in Sec.~\ref{Ch5:background}, to express the cost for an individual data point as $(\lambda,\sum_{i=1}^{3}2\lambda/\sigma^{2})$-RDP. And, the privacy cost for a batch of data points $\{x_i,..,x_B\}$ can be similarly extended to be   $(\lambda,\sum_{i=1}^{B}\sum_{i=1}^{3}2\lambda/\sigma^{2})$ or equivalently $(\lambda,6B\lambda/\sigma^{2})$. \tiny\qedsymbol

\normalsize Next, to amplify the privacy protection for the generator, theorem 2.2.3 defined in Sec.~\ref{Ch5:background} is analogously used. However, in this case, the original training dataset is divided into disjoint subsets of equal size where a unique discriminator is trained for each subset independently. The size of each subsampled data is defined as $N_{d}/N$ where $N_{d}$ is the total number of discriminators and $N$ is the size of the full training dataset. Thus, during training, one out of the total number of discriminators is chosen randomly for every iteration to provide gradient updates to the generator on the basis of it's corresponding subsampled dataset. In this way, the subsampling rate for the generator is defined to be $\gamma = 1/N_{d}$.

Unfortunately, training multiple discriminators on smaller subsamples is problematic due to the lack of enough training iterations for any given discriminator in comparison to the generator. Moreover, reducing the number of samples via subsampling increases the potential of over-fitting  the discriminators on it's respective subsample. \cite{chen2020gs} recommends to alleviate the first problem by pre-training the multiple discriminator networks with a standard generator without DP. Since, pre-training the discriminators reliably doesn't breach the DP guarantees for the generator. However, in practice, the results were not found to be affected by the presence of pre-trained discriminators in preliminary experiments.

Lastly, definition 2.2.2 defined in~Sec.\ref{Ch5:background} is used to convert the overall cumulative privacy cost computed in terms of RDP back to $(\epsilon,\delta)$-DP for both approaches. Practically, these computations are performed via the official implementation\footnote{\url{https://github.com/yuxiangw/autodp}} provided by \cite{wang2019subsampled}.

\section{Experimental Analysis}
\label{Ch5:EA}

\subsection{Experimental Setup}
\label{Ch5:ES}

\textbf{Datasets}- To evaluate DP-CTABGAN, 3 out of the 5 datasets introduced in Sec.~\ref{Ch3:DD} are used i.e., Adult~\cite{UCIdataset}, Credit~\cite{kagglecredit} and Loan~\cite{kaggleloan}. Refer to Tab.~\ref{table:DDE} detailing each dataset. 
\\
\textbf{Baselines}- Both variants of DP-CTABGAN are compared with 2 state-of-the-art architectures: PATE-GAN~\cite{pategan}\footnote{\url{https://github.com/vanderschaarlab/mlforhealthlabpub/tree/main/alg/pategan}} and DP-WGAN~\cite{xie2018differentially}\footnote{\url{https://github.com/BorealisAI/private-data-generation/blob/master/models/dp_wgan.py}}. Additionally, to present a fair comparison between DP-WGAN and PATE-GAN, a common network architecture for the both the generator and discriminator is used (refer to Sec.~\ref{appendix:1}). Tab.~\ref{Ch5:tab1}. outlines the salient features of all methods used in this evaluation. 

Lastly, it is important to note that for DP-WGAN, the authors originally derive the privacy cost using the moment accountant technique~\cite{abadi2016deep}. However, in this work, to compare fairly across different approaches that all making use of DP-SGD with gaussian mechanisms, the more optimal subsampled RDP accountant~\cite{mironov2017renyi,wang2019subsampled} is used. This is because, the RDP-accountant allows for even tighter bounds on the privacy budget than the moment accountant enabling less noise to be added during training for ensuring similar privacy guarantees. 

\subsection{Evaluation Metrics}
\textbf{Statistical Similarity $\&$ ML Utility}- The evaluation metrics concerning the statistical similarity and ML utility is borrowed from Sec.~\ref{Ch3:metrics}. However, there are a few notable differences worth mentioning. 

Firstly, with respect to the statistical similarity, unlike previous chapters, the WD is calculated after performing a min-max normalisation\footnote{\url{https://scikit-learn.org/stable/modules/generated/sklearn.preprocessing.MinMaxScaler.html}} for both the real and synthetic values using the real maximum and minimum values corresponding to a particular column. This is done for averaging the wasserstein distances across columns with drastically varying scales more reliably. 

Secondly, for evaluating the ML utility, the average precision score (APR)\footnote{https://scikit-learn.org/stable/modules/generated/sklearn.metrics.average\_precision\_score.html} is introduced to provide a reliable source of performance in comparison to the AUC given the imbalance in datasets used. Moreover, the SVM model is eliminated from the study due to practical limitations with outputting predicted probabilities in a time-efficient manner. Lastly, the Min-Max normalisation is used as a pre-processing step before training of ML models as used in the evaluation done by \cite{pategan}. \\
\\
\textbf{Inference Attacks}- This chapter introduces two new metrics for evaluating the empirical robustness of GANs against malicious privacy attacks. More specifically, the membership and attribute inference attacks are launched against each model to expose the risk of privacy loss based on the rigorous framework provided by \cite{priv_mirage}.\\
\\
\textbf{The membership inference attack~\cite{chen2020gan}} is a binary classification problem in which an attacker tries to predict if a particular target data point $t$ has been used to train a victim generative model. This work assumes that the attacker only needs access to a black-box tabular GAN model, a reference dataset $\mathcal{R}$ and $t$ for which the inference must be made~\cite{priv_mirage}. 

\begin{figure}[htb]
    \centering
    \includegraphics[scale=.06]{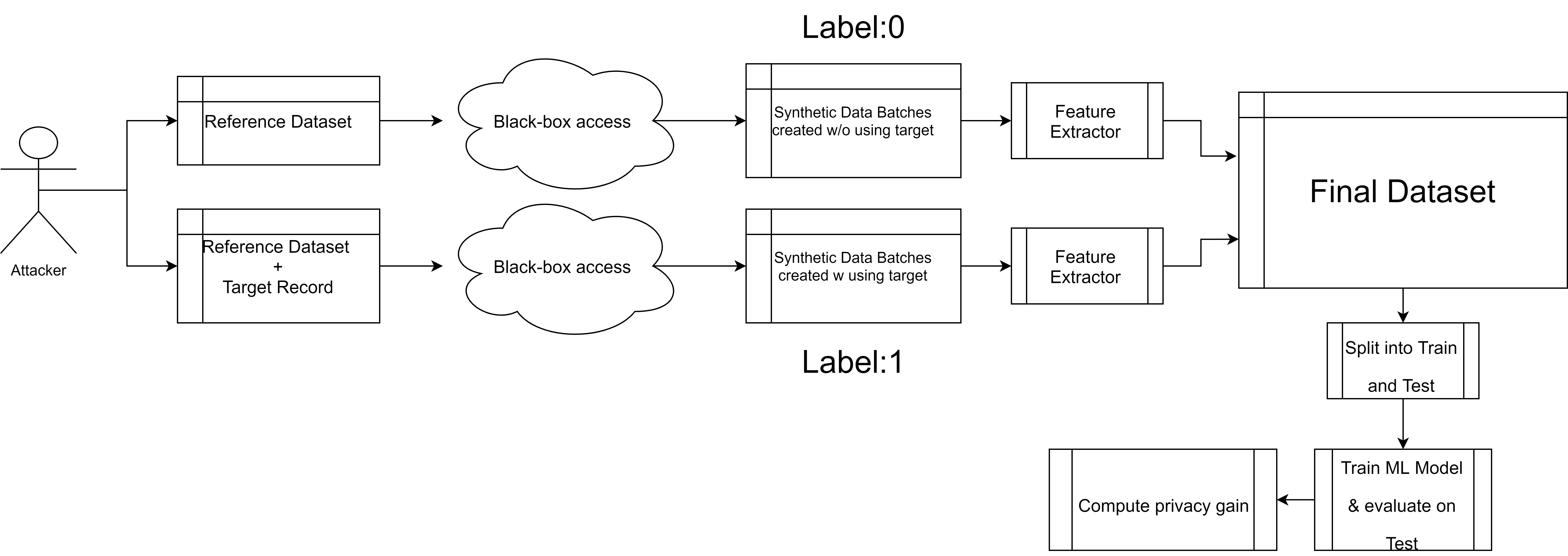}
    \caption{\centering Membership Inference Attack Pipeline}
    \label{fig:MIE}
\end{figure}

As illustrated in Fig.~\ref{fig:MIE}, to launch an attack, the attacker prepares two training datasets with and without the target record $t$ using the reference dataset $\mathcal{R}$ (i.e., $\mathcal{R}$, $\mathcal{R}\oplus t$). Next, the attacker uses black-box access to the model for training two separate models on each dataset. The attacker then uses these to generate $s$ batches of synthetic data each consisting of $r$ rows, represented as $\mathcal{S}^{s}_{r}$. The synthetic batches are assigned a label of 0 and 1, respectively, based on the presence of $t$ in the training dataset. 

Thereafter, each batch of synthetic data is processed by a feature extraction method summarizing the information contained in each batch into a single vector. This is done in two ways: (i) naive extraction- computes the mean, median, and variance of every continuous column and the length of unique categories as well as the most and least frequently occurring category for every categorical column (ii) correlation extraction- computes the pairwise correlations between all columns where the categorical columns are dummy-encoded. 

This leads to the creation of a final dataset, containing an equal number of processed samples. This is split into train and test datasets. An attack model is trained on the training dataset and used to compute the privacy gain as $P_{Gain}=\frac{(P_{Real} - P_{Fake})}{2}$ where $P_{Fake}$ is the attack model's average probability of successfully predicting the correct label in the test-set and $P_{Real}=1$ since having access to the original training data ensures full knowledge of $t$'s presence~\cite{priv_mirage}. 

To conduct the membership inference evaluation, 4000 rows of real data were sampled from each dataset to form the reference dataset (i.e., $\mathcal{R}$) to train the synthetic models. Each batch for feature extraction was chosen to be of size $r=400$. And, $s=1200$ batches were generated such that the training dataset was of size 1000 with balanced number of classes. And, the test set contained 200 samples with balanced classes. To train the attack model, the Random-Forest-Classifier\footnote{https://scikit-learn.org/stable/modules/generated/sklearn.ensemble.RandomForestClassifier.html} was used. The experiments were repeated 5 times with 5 different target records $t$ for each dataset and the results were averaged.
\\
\\
\textbf{An attribute inference attack~\cite{priv_mirage}} is defined as a regression problem where the attacker attempts to predict the values of a sensitive target column provided he/she has black-box access to a generative model.

\begin{figure}[htb]
    \centering
    \includegraphics[scale=.062]{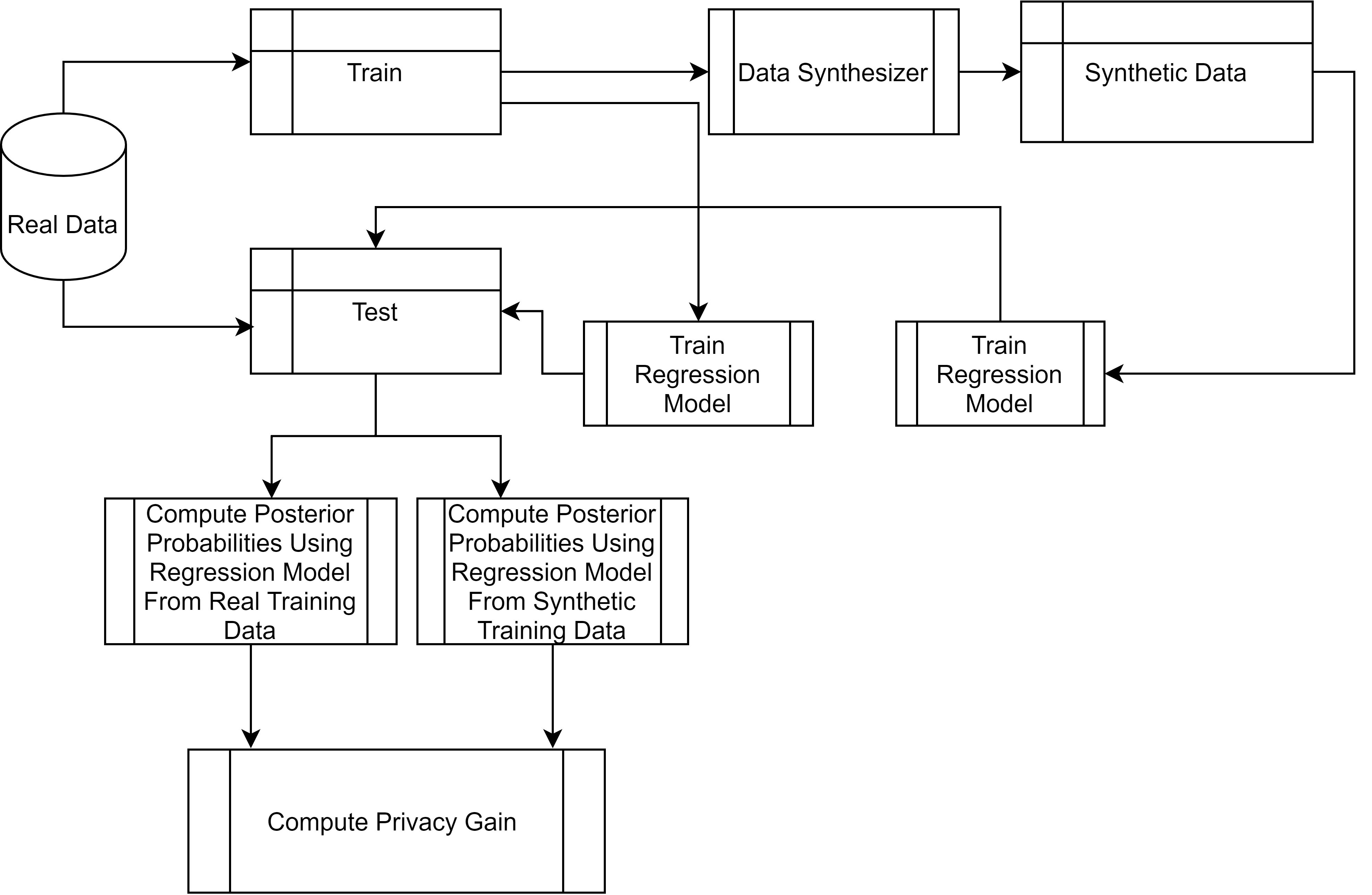}
    \caption{\centering Attribute Inference Attack Pipeline}
    \label{fig:MIA1}
\end{figure}

To launch an attribute inference attack and evaluate the privacy risk (refer to Fig.~\ref{fig:MIA1}), a dataset $\mathcal{R}$ sampled from the real distribution is split into train and test datasets, respectively (i.e., $\mathcal{R}_{Train}$ $\&$ $\mathcal{R}_{Test}$). $\mathcal{R}_{Train}$ is fed into a generative model for generating a corresponding synthetic training dataset (i.e., $\mathcal{G}_{Train}$).

A linear regression model\footnote{https://scikit-learn.org/stable/modules/generated/sklearn.linear\_model.LinearRegression.html} is then used for estimating the relationship between the independent variables known to the attacker and the dependent sensitive variable for both $\mathcal{R}_{Train}$ and $\mathcal{G}_{Train}$. Then, to evaluate the privacy risk, the privacy gain is computed similarly as: $P_{Gain} = \frac{(P_{Real}-P_{Fake})}{2}$, where $P_{Real}$ $\&$ $P_{Fake}$ denote the average posterior probabilities of correctly predicting the sensitive attribute on the real testset given the linear models fitted on $\mathcal{R}_{Train}$ and $\mathcal{G}_{Train}$, respectively~\cite{priv_mirage}.  

For performing the attribute inference evaluation, 5000 real samples (i.e.,$\mathcal{R}$) were sampled from each dataset where 4900 samples were used for creating the training dataset (i.e., $\mathcal{R}_{Train}$) and 100 for the testing dataset (i.e.,$\mathcal{R}_{Test}$). Moreover, the sensitive attribute for the datasets Adult, Loan and Credit were chosen to be "Age","Age" and "Amount", respectively. The experiment was repeated 5 times and the average results are presented.

\subsection{Results}
 
 This section presents the results for all baselines based on the criteria established previously. Note that for measuring the statistical similarity and ML efficacy, the privacy budget $\epsilon$ is varied between 1 and 100 to study the influence of a strong vs weak privacy constraint, respectively. 
 
 However, for evaluating the risk of privacy loss via membership and attribute inference attacks, a strict privacy budget of $\epsilon = 1$ is chosen as commonly used in prior work~\cite{pategan}. This is done to thoroughly test the effectiveness of DP techniques offering strong theoretical guarantees empirically. Refer to Sec.~\ref{appendix:2} of the appendix for details concerning hyper-parameters used to generate samples from all baselines for conducting the experiments.
 
Lastly, all result tables feature DP-CTABGAN with no privacy budget (i.e.,$\epsilon=\infty$) simply denoted as CTAB-GAN to be used as a reference point for examining the influence of differential privacy for training CTAB-GAN. Note that $\delta=1e-5$ is fixed across all experiments and the best results are highlighted in bold among only those models that are trained with finite privacy budgets.\\
\\
\textbf{Statistical Similarity $\&$ ML Utility}-
\label{Ch5:Res}
\begin{enumerate}

    \item \textbf{Statistical Similarity}- As shown in Tab.~\ref{table:SS_allE1} and Tab.~\ref{table:SS_allE2}, among all baseline models, D-DP-CTABGAN is the only model which consistently improves across all three metrics when the privacy budget is increased. 

Similarly, G-DP-CTABGAN sees an improvement across both the Avg-JSD and Avg-WD. However, PATE-GAN and DP-WGAN do not show signs of improvement consistently across any of the metrics. Moreover, they perform worse than both variants of DP-CTABGAN at both levels of epsilon. 

This highlights their inability to capture the statistical distributions during training despite a loose privacy budget purely due to the lack of an effective training framework. 

Lastly, it is worth noting that G-DP-CTABGAN features the best correlation distance at $\epsilon=1$ and $\epsilon=100$ showcasing that training the discriminator reliably is hugely beneficial for capturing correlations in the data as compared to D-DP-CTABGAN. Naturally, there is still a huge performance gap between CTAB-GAN and both variants of DP-CTABGAN due to the application of DP.\\

\begin{table}[htb]
\centering
\caption{\centering Statistical similarity: 3 measures averaged over 3 datasets with a privacy budget of $\epsilon=1$}
\begin{tabular}{|c|c|c|c|}
\hline
\textbf{Method} & \textbf{Avg JSD} & \textbf{Avg NWD}  & \textbf{Diff. Corr.} \\
\hline
\small{PATE-GAN}    & 0.487   & 0.259   & 3.982  \\
\small{DP-WGAN} &0.299   &  0.232 & 3.834 \\
\small{D-DP-CTABGAN}   & \textbf{0.246}  & \textbf{0.063}  &  4.168 \\
\small{G-DP-CTABGAN}   &  0.376  &   0.189 &   \textbf{3.065}   \\
\small{CTAB-GAN}   & 0.028  &  0.01 & 1.607 \\
\hline
\end{tabular}
\label{table:SS_allE1}
\end{table}

\begin{table}[htb]
\centering
\caption{\centering Statistical similarity: 3 measures averaged over 3 datasets with a privacy budget of $\epsilon=100$ }
\begin{tabular}{|c|c|c|c|}
\hline
\textbf{Method} & \textbf{Avg JSD} & \textbf{Avg NWD}  & \textbf{Diff. Corr.} \\
\hline
\small{PATE-GAN}    &  0.358  & 0.259   & 4.837  \\
\small{DP-WGAN} & 0.304 &  0.222   & 4.57 \\
\small{D-DP-CTABGAN}   & \textbf{0.127} &  \textbf{0.047} &  3.648 \\
\small{G-DP-CTABGAN}   &  0.389 & 0.174 &   \textbf{3.21}   \\
\small{CTAB-GAN}   & 0.028  &  0.01 &   1.607  \\
\hline
\end{tabular}
\label{table:SS_allE2}
\end{table}

\item \textbf{ML Efficacy}- From the results presented in Tab.~\ref{table:ML_allE1} and Tab.~\ref{table:ML_allE2}, surprisingly PATE-GAN performs worse in terms of ML utility with a looser privacy budget. This is mainly because the student discriminator is trained solely with generated samples of poor statistical similarity as found in Tab.~\ref{table:SS_allE2}. 

Moreover, it is found similar as before that only the D-DP-CTABGAN model consistently improves across all metrics with a looser privacy budget. And, showcases the best performance for both the f1-score and APR metrics with different privacy budgets across all baselines. This finding suggests that the based on the implementations of D-DP-CTABGAN and G-DP-CTABGAN used in this work, training the discriminator with DP guarantees is more optimal. This is in line with the challenges faced by G-DP-CTABGAN due to subsampling which hugely degrades performance by training multiple discriminators each using a smaller number of samples. 

Finally, the performance increase of D-DP-CTABGAN in comparison to other baselines can be explained by it's sophisticated neural network architecture (i.e., conditional GAN) and improved training objective (i.e., wasserstein loss with gradient penalty). However, as a consequence of the application of DP, the performance decrease in comparison to CTAB-GAN is noticeably large.\\

\begin{table}[htb]
\centering
\caption{\centering Difference of ML accuracy (\%), F1-score, AUC and APR between original and synthetic data: average over 3 different datasets and a privacy budget $\epsilon=1$}
\begin{tabular}{|c|c|c|c|c|}
\hline
\textbf{Method} & \textbf{Accuracy} & \textbf{AUC}  & \textbf{APR}& \textbf{F1-Score} \\
\hline
\small{PATE-GAN}    & 10.8\% & \textbf{0.246} &   0.576 & 0.367 \\
\small{DP-WGAN}     & \textbf{8.2\%}  & 0.408 &   0.58 & 0.368 \\
\small{D-DP-CTABGAN}    & 16.1\%&  0.302& \textbf{0.483} & \textbf{0.34}\\
\small{G-DP-CTABGAN}    &  32.3\%& 0.377 &  0.604 & 0.454 \\
\small{CTABGAN}    & 2.6\% & 0.042 &  0.143 & 0.097 \\
\hline
\end{tabular}
\label{table:ML_allE1}
\end{table}

\begin{table}[htb]
\centering
\caption{\centering Difference of ML accuracy (\%), F1-score, AUC and APR between original and synthetic data: average over 3 different datasets and a privacy budget $\epsilon=100$}
\begin{tabular}{|c|c|c|c|c|}
\hline
\textbf{Method} & \textbf{Accuracy} & \textbf{AUC}  & \textbf{APR}& \textbf{F1-Score} \\
\hline
\small{PATE-GAN}    & 37.4\% & 0.416 &   0.566 & 0.412     \\
\small{DP-WGAN}     & \textbf{10.8\%}  & 0.373 & 0.592 & 0.364 \\
\small{D-DP-CTABGAN}    & 13\%    &  \textbf{0.265} & \textbf{0.475} &  \textbf{0.262}  \\
\small{G-DP-CTABGAN}    &  13.7\% & 0.387  &  0.565 & 0.374 \\
\small{CTABGAN}    & 2.6\% & 0.042 &  0.143 & 0.097 \\
\hline
\end{tabular}
\label{table:ML_allE2}
\end{table}
\end{enumerate}
\textbf{Privacy Impact Against Inference Attacks}-

\begin{enumerate}
    \item \textbf{Membership Inference Attack}- From the results shown in Tab.~\ref{table:PP_allE}, it is found that all DP baselines provide an empirical privacy gain close to 0.25 for both feature extraction methods. This indicates that differential private methods provide a strong privacy protection against membership attacks. And ensures that the average probability of success for any attack is close to the attacker's original prior i.e 0.5. Furthermore, it is found that D-DP-CTABGAN and G-DP-CTABGAN provide the highest security against a membership attack with naive and correlation feature extraction methods, respectively. 
    
    Moreover, there is a clear decrease in the privacy gain achieved by CTABGAN showcasing that DP is needed to provide a stronger defense against membership inference attacks. 
    
    
    \item \textbf{Attribute Inference Attack}- Tab.~\ref{table:PP_allE} shows that PATE-GAN provides the greatest security. Moreover, both versions of DP-CTABGAN provide a lesser privacy protection than other baselines. And, CTABGAN provides the worst security. This is due to the superior quality of the synthetic data offered by CTABGAN and it's DP variants which enhances the attacker's probability of successfully inferring sensitive information. These results highlight the inherent trade-off between privacy and data utility i.e., increasing the utility directly worsens the privacy and vice versa. 

    It is worth noting that the privacy gain for attribute inference attack for all baselines is close to 0 suggesting that the overall privacy protection offered against attribute inference attacks is quite low. However, it should be noted that the privacy gain is computed with respect to the real data. Thus, in case the real data itself provides a low probability of successfully inferring the correct target values for a sensitive attribute, then the synthetic dataset will perform in a similar manner resulting in a privacy gain close to 0. 

\end{enumerate}

\begin{table}[htb]
\centering
\caption{\centering Empirical privacy gain against membership attack with naive $\&$ correlation feature extraction and attribute inference attack: average over 3 different datasets with a privacy budget $\epsilon=1$} 
\resizebox{0.8\columnwidth}{!}{
\begin{tabular}{|c|c|c|c|}

\hline
\textbf{Method} & \textbf{Naive Privacy Gain} & \textbf{Correlation Privacy Gain} & \textbf{Attribute Inference Privacy Gain}\\
\hline
\small{PATE-GAN}    &  0.25   & 0.25 & \textbf{0.042} \\
\small{DP-WGAN}     &  0.255  & 0.256 & 0.04 \\
\small{D-DP-CTABGAN}&  \textbf{0.266}  & 0.248 & 0.037 \\
\small{G-DP-CTABGAN}&  0.245  & \textbf{0.26} & 0.038  \\
\small{CTABGAN}&  0.238  & 0.233 & -2e-4  \\
\hline
\end{tabular}
}
\label{table:PP_allE}
\end{table}

\section{Conclusion}
\label{Ch5:Conclusion}

In this chapter, two variants of DP-CTABGAN were proposed and their corresponding privacy analyses were underlined. Based on theoretical derivations and empirical results, D-DP-CTABGAN resulted in a superior configuration for integrating DP guarantees into CTAB-GAN. Moreover, D-DP-CTABGAN consistently outperformed existing state-of-the-art baselines concerning generated sample quality in terms of both statistical similarity and ML utility metrics. 

Additionally, both variants of DP-CTABGAN were found to be resilient towards membership and attribute inference attacks. Therefore, this work showcases the effectiveness of DP for protecting the privacy of sensitive datasets being used for training tabular GANs.

However, further enhancement of the quality of synthetic data at strict privacy budgets (i.e., $\epsilon \leq 1$) is still needed. Ultimately, there is an inherent trade-off between privacy and utility and obtaining the most optimal balance between both is left for future work.  


%% file: Content/Chapters/6_Conclusion.tex
\chapter{Conclusion}\label{ch6}

Tabular data is a key asset for data-driven industries that are fueled by modern advancements in the field of machine learning. However, utilising real tabular data risks leaking private information about individuals. Therefore, tabular GANs have gained vital importance as a viable solution to utilise tabular data without breaching privacy. 
\\
This thesis dealt with three main research questions pertaining to tabular GANs:
\begin{itemize}
    \item \textit{"What are the performance capabilities of existing tabular GANs?"}- To answer this research question, 4 state-of-the-art tabular GAN models were extensively evaluated on 5 datasets in terms of their ML utility, statistical similarity and privacy. And their major strengths and weaknesses were highlighted.
    
    \item \textit{"How to improve upon the tabular generation quality of state-of-the-art tabular GANs?"}- Based on the exposed difficulties of existing methods, this work developed a novel conditional tabular GAN architecture, CTAB-GAN. CTAB-GAN was shown to effectively handle "mixed" data types and skewed variables. And, improved upon prior work in data utility for ML applications by up to 17\% in accuracy for 5 ML models on complex datasets while maintaining a safer privacy distance than prior-work.
    
    \item \textit{"How to prevent privacy leakage for tabular GANs?"}- The use of differential privacy for enhancing the privacy of tabular GAN training was examined.  Moreover, CTABGAN with DP guarantees was rigorously tested along side state-of-the-art DP-GANs with respect to generation quality and privacy protection against membership and attribute inference attacks. Our results using 3 datasets and 4 ML models showed that DP-CTABGAN maintains the highest data utility by up to 18\% in terms of the average precision score as compared to prior work while reliably withstanding privacy attacks.   
\end{itemize}

To conclude the thesis, a few important limitations of this work and corresponding future directions are highlighted: 
\begin{itemize}
    \item CTAB-GAN makes use of convolution operations that rely on a square matrix representation of the input data. This requires additional padding that adds useless information to the data. Therefore, the use of rectangular kernel operations that can be executed directly on rectangular shaped input data can be further looked into. 
    \item CTAB-GAN's suffers from poor convergence on small sized datasets. Therefore, effectively reducing the training complexity of CTAB-GAN for smaller datasets is needed. And so, simpler data transforms that can allow to learn dependencies between variables without increasing the input dimensionality needs further exploration.
    \item There is large gap between the data utility of synthetic data generated with and without using strict privacy guarantees. Moreover, determining the most optimal privacy budget $\epsilon$ that best balances the privacy/utility trade off requires future consideration.
    
\end{itemize}

\afterpage{\blankpage}

%% file: Content/appendix.tex
\chapter{Differential Privacy Experimental Setup}

The supplementary material highlights the network architecture shared between PATE-GAN~\cite{pategan} and DP-WGAN~\cite{xie2018differentially} as mentioned in~Sec.~\ref{Ch5:ES}. Additionally, it provides hyper-parameters used for conducting the data utility (i.e., statistical similarity $\&$ ML utility) as well as the membership and attribute inference attack experiments.

\section{Network Architecture}
\label{appendix:1}

The network architecture for training PATE-GAN is used identically to their original implementation provided on github\footnote{\url{https://github.com/vanderschaarlab/mlforhealthlabpub/tree/main/alg/pategan}}. And, the network structure of DP-WGAN\footnote{\url{https://github.com/BorealisAI/private-data-generation/blob/master/models/dp_wgan.py}} used in the experiments has been modified from the original to have the exact neural network architecture for the discriminator and generator networks as that of PATE-GAN. This is done to study the performance of DP-WGAN in relation to PATE-GAN.

The generator network of PATE-GAN comprises of a shallow neural network with 3 fully connected layers that each comprise of $4*l$ nodes where $l$ is the length of each row in the original data. The first 2 fully connected layers are followed by a \textit{Tanh activation} whereas for the last layer a \textit{Sigmoid activation} is used. This is done to bring the values generated in the range of [0,1] which is the same range as the normalised data used for training. 

The student discriminator network of PATE-GAN comprises of a shallow neural network with 2 fully connected layers with $l$ nodes. The first layer is followed by a \textit{ReLU activation function} whereas the output of the second layer is used directly for computing the KL divergence loss of the discriminator as shown in Eq.~\ref{eq:gan}. 

\section{Network Hyper-parameters}
\label{appendix:2}

Across all baselines, the batch size was set to 64. Moreover, for PATE-GAN and DP-WGAN, default hyper-parameters as found in the code-bases were utilized. Thus, PATE-GAN uses 10 as the default number of teacher discriminators for all experiments. And DP-WGAN, uses [-0.01,0.01] to clamp the weights of the discriminator and $0.1$ as the gradient norm bound $C$.   

Additionally, Tab.~\ref{tab:app0} and Tab.~\ref{tab:app1} provide details concerning the differential-private hyper-parameters such as the noise scale used and the number of training epochs\footnote{Note that in the original implementation of PATE-GAN, the privacy budget $\epsilon=1$ is expended with just one iteration over a single batch. Therefore, in the epochs columns, the number of iterations over a single batch is displayed.} required for generating synthetic tabular data with the corresponding privacy budget epsilon (i.e., $\epsilon$) to conduct the data utility experiments and privacy attack experiments in Sec.~\ref{Ch5:ES}.

\begin{table}[htb]
\centering
\caption{\centering Differential privacy hyper-parameters for conducting statistical similarity and ML utility experiments.}
\resizebox{0.8\columnwidth}{!}{
\begin{tabular}{|c|c|c|c|c|c|}
\hline
\textbf{Model}  & \textbf{Dataset} & \textbf{No. of Discriminators} & \textbf{Noise Scale} & \textbf{Epochs} & \textbf{Epsilon} \\
\hline
PATE-GAN     & Adult   & 1                        & 1           & 1      & 1       \\
PATE-GAN     & Credit  & 1                        & 1           & 1      & 1       \\
PATE-GAN     & Loan    & 1                        & 1           & 1      & 1       \\
DP-WGAN      & Adult   & 1                        & 1.012       & 1      & 1       \\
DP-WGAN      & Credit  & 1                        & 1.012       & 1      & 1       \\
DP-WGAN      & Loan    & 1                        & 1.33        & 1      & 1       \\
D-DP-CTABGAN & Adult   & 1                        & 1.06        & 1      & 1       \\
D-DP-CTABGAN & Credit  & 1                        & 1.06        & 1      & 1       \\
D-DP-CTABGAN & Loan    & 1                        & 1.58        & 1      & 1       \\
G-DP-CTABGAN & Adult   & 1000                     & 3.518       & 1      & 1       \\
G-DP-CTABGAN & Credit  & 1000                     & 3.53        & 1      & 1       \\
G-DP-CTABGAN & Loan    & 1000                     & 1.28        & 1      & 1       \\
PATE-GAN     & Adult   & 1                        & 1           & 795    & 100     \\
PATE-GAN     & Credit  & 1                        & 1           & 795    & 100     \\
PATE-GAN     & Loan    & 1                        & 1           & 795    & 100     \\
DP-WGAN      & Adult   & 1                        & 0.33        & 6      & 100     \\
DP-WGAN      & Credit  & 1                        & 0.33        & 6      & 100     \\
DP-WGAN      & Loan    & 1                        & 0.38        & 7      & 100     \\
D-DP-CTABGAN & Adult   & 1                        & 0.36        & 5      & 100     \\
D-DP-CTABGAN & Credit  & 1                        & 0.36        & 5      & 100     \\
D-DP-CTABGAN & Loan    & 1                        & 0.42        & 4      & 100     \\
G-DP-CTABGAN & Adult   & 50                       & 0.867       & 1      & 100     \\
G-DP-CTABGAN & Credit  & 100                      & 0.874       & 1      & 100     \\
G-DP-CTABGAN & Loan    & 100                      & 1.089       & 4      & 100    \\
\hline
\end{tabular}
}
\label{tab:app0}
\end{table}

\begin{table}[htb]
\centering
\caption{\centering Differential privacy hyper-parameters for conducting membership and attribute inference attacks.}
\resizebox{0.8\columnwidth}{!}{
\begin{tabular}{|c|c|c|c|c|c|c|}
\hline
\textbf{Model} & \textbf{Dataset} & \textbf{No of Discriminators} & \textbf{Noise Scale (Membership)} & \textbf{Noise Scale (Attribute)} & \textbf{Epochs} & \textbf{Epsilon} \\ \hline
PATE-GAN     & Adult  & 1    & 1    & 1    & 1 & 1 \\
PATE-GAN     & Credit & 1    & 1    & 1    & 1 & 1 \\
PATE-GAN     & Loan   & 1    & 1    & 1    & 1 & 1 \\
DP-WGAN      & Adult  & 1    & 1.33 & 1.25 & 1 & 1 \\
DP-WGAN      & Credit & 1    & 1.33 & 1.25 & 1 & 1 \\
DP-WGAN      & Loan   & 1    & 1.33 & 1.25 & 1 & 1 \\
D-DP-CTABGAN & Adult  & 1    & 1.67 & 1.56 & 1 & 1 \\
D-DP-CTABGAN & Credit & 1    & 1.67 & 1.56 & 1 & 1 \\
D-DP-CTABGAN & Loan   & 1    & 1.67 & 1.56 & 1 & 1 \\
G-DP-CTABGAN & Adult  & 1000 & 1.28 & 1.37 & 1 & 1 \\
G-DP-CTABGAN & Credit & 1000 & 1.28 & 1.37 & 1 & 1 \\
G-DP-CTABGAN & Loan   & 1000 & 1.28 & 1.37 & 1 & 1 \\
\hline
\end{tabular}
}
\label{tab:app1}
\end{table}